\documentclass[journal,twoside,web]{ieeecolor}
\pdfoutput=1
\usepackage{tmi}
\usepackage{cite}
\usepackage{amsmath,amssymb,amsfonts}
\usepackage{mathrsfs}
\usepackage{algorithm,algpseudocode}
\usepackage{eucal}
\usepackage{bbm}
\usepackage{bm}
\usepackage{multicol}
\usepackage{multirow}
\usepackage{subfigure}
\usepackage[dvipsnames]{xcolor}
\usepackage[hidelinks]{hyperref}
\usepackage{colortbl}
\usepackage{arydshln}
\usepackage{bbding}
\usepackage{graphicx}
\usepackage{textcomp}
\usepackage{algorithmicx}
\usepackage{algpseudocode}
\usepackage{float}

\newcommand{\revised}[1]{{\color[rgb]{0,0,0.0}{#1}}}
\newcommand{\xp}[1]{{\color[rgb]{0.9,0.1,0.1}{[XP:#1]}}}
\makeatletter
\DeclareRobustCommand\onedot{\futurelet\@let@token\@onedot}
\def\@onedot{\ifx\@let@token.\else.\null\fi\xspace}

\makeatother

\algnewcommand{\LineComment}[1]{\State \(\triangleright\) #1}
\def\BibTeX{{\rm B\kern-.05em{\sc i\kern-.025em b}\kern-.08em
    T\kern-.1667em\lower.7ex\hbox{E}\kern-.125emX}}
\begin{document}

\title{Less is More: Surgical Phase Recognition from Timestamp Supervision}

\author{Xinpeng Ding, Xinjian Yan, Zixun Wang, Wei Zhao, Jian Zhuang, Xiaowei Xu and Xiaomeng Li
\thanks{Manuscript received XX XX, 2021. X. Ding, Z. Wang and X. Li are with the Department of Electronic and Computer Engineering, The Hong Kong University of Science and Technology, Hong Kong SAR, China. W. Zhao is with the School of Physics, Beihang University, Beijing, China, and also with Beihang Hangzhou Innovation Institute Yuhang
Xixi Octagon City, Yuhang District, Hangzhou, China (email: craddywagn@gmail.com; xpding.xidian@gmail.com; zhaow20@buaa.edu.cn; eexmli@ust.hk)}
\thanks{Copyright (c) 2021 IEEE. Personal use of this material is permitted. Permission from IEEE must be obtained for all other uses, including reprinting/republishing this material for advertising or promotional purposes, collecting new collected works for resale or redistribution to servers or lists, or reuse of any copyrighted component of this work in other works.}}

\maketitle
 
\begin{abstract}
Surgical phase recognition is a fundamental task in computer-assisted surgery systems.
\revised{
Most existing works are under the supervision of expensive and time-consuming full annotations, which require the surgeons to repeat watching videos to find the precise start and end time for a surgical phase.
}
{In this paper, we introduce timestamp supervision for surgical phase recognition to train the models with timestamp annotations, where the surgeons are asked to identify only a single timestamp within the temporal boundary of a phase.
This annotation can significantly reduce the manual annotation cost compared to the full annotations.
}
%
%
{To make full use of such timestamp supervisions, we propose a novel method called uncertainty-aware temporal diffusion (UATD) to generate trustworthy pseudo labels for training.}
{
Our proposed UATD is motivated by the property of surgical videos,~\emph{i.e.}, the phases are long events consisting of consecutive frames.
To be specific, UATD diffuses the single labelled timestamp to its corresponding high confident (~\emph{i.e.}, low uncertainty) neighbour frames in an iterative way.
}
\revised{Our study uncovers unique insights of surgical phase recognition with timestamp supervisions:
1) timestamp annotation can reduce $74\%$ annotation time compared with the full annotation, and surgeons tend to annotate those timestamps near the middle of phases;
2) extensive experiments demonstrate that our method can achieve competitive results compared with full supervision methods, while reducing manual annotation cost;
3) less is more in surgical phase recognition,~\emph{i.e.}, less but discriminative pseudo labels outperform full but containing ambiguous frames;
4) the proposed UATD can be used as a plug and play method to clean ambiguous labels near boundaries between phases, and improve the performance of the current surgical phase recognition methods.
Code and annotations obtained from surgeons are available at \url{https://github.com/xmed-lab/TimeStamp-Surgical}.
}
%
%
%
%
%
%
\end{abstract}


\section{Introduction}
\label{sec:introduction}
\begin{figure}[t]
\centering 
\label{fig:timestamp}
\includegraphics[width=0.48\textwidth,height=0.4\textheight]{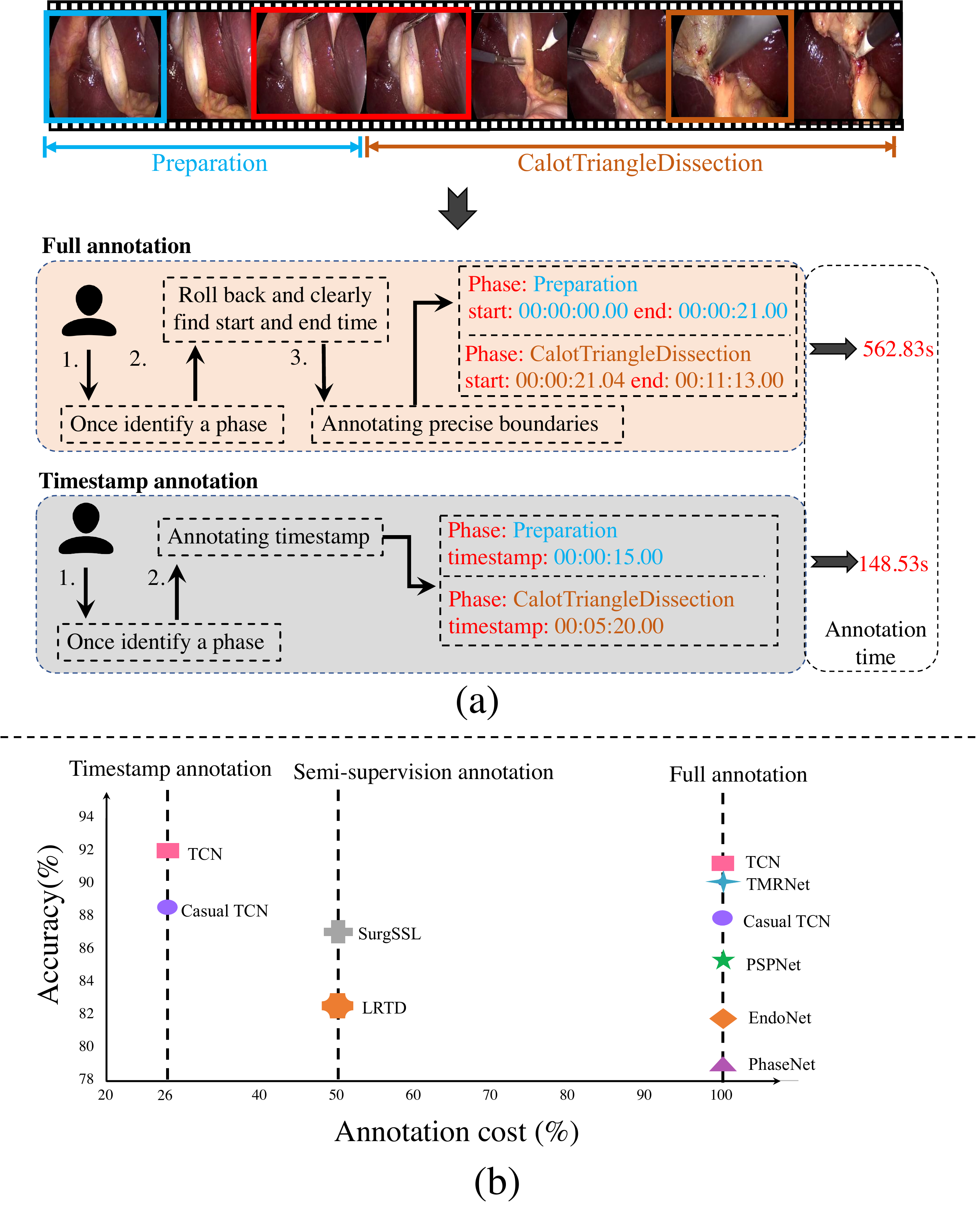}
\caption{(a)~\revised{Comparison of the full annotation and our proposed timestamp annotation. When labelling a phase in full annotation, the annotator needs to roll back and find the precise start and end time. In our timestamp annotation, only a single timestamp is labelled without identifying the start and end time, which can save annotation cost and is much faster than the full annotation.
We invite two surgeons to conduct full and timestamp annotations, and record their annotation times.
We finally observe that they took an average of $562.83$ and $148.53$ seconds per video for full annotation and timestamp annotation, respectively.
}
(b) \revised{The trade-off between manual annotation cost and accuracy for different methods.
Compared with existing methods, our method achieves the competitive performance while using only $26\%$ manual annotation cost compared with the full supervision.
}}
\end{figure}
\IEEEPARstart{C}{omputer-assisted} surgery systems can improve the surgery's quality and ensure the patients' safety in modern operating rooms~\cite{maier2017surgical,moglia2016systematic}. 
Surgical phase recognition is one key component of computer-assisted surgery systems, which aims to predict which phase is occurring at the current frame~\cite{jin2017sv,jin2021temporal}.
It can be used for automatic indexing of surgical video databases~\cite{twinanda2016endonet}, monitoring surgical process~\cite{bricon2007context}, scheduling surgeons~\cite{bhatia2007real} and assessing surgeons' skills~\cite{liu2021towards}.
In recent years, automated surgical phase recognition has featured deep learning~\cite{he2016deep,simonyan2014very,liu2021swin} and has reached promising recognition performance~\cite{ding2022exploring,twinanda2016endonet,gao2021trans}.
Most current surgical phase recognition approaches require full annotations from surgeons,~\emph{i.e.}, the surgeons need to find the precise start and end time for a surgical phase.
To this end, the surgeon should repeat watching the video at a very slow speed to find a specific time for the start of the phase. Then, the surgeon needs to continue to watch the video and find the precise end time of the phase. 
As shown in Fig.~\ref{fig:timestamp}~(a), this full annotation is very time-consuming,~\emph{e.g.}, surgeons need to spend an average of 562.83 seconds to annotate a video.
Furthermore, the boundaries between different phases are usually ambiguous~\cite{ding2022exploring}.
\revised{
Due to the subjective of different surgeons, they would provide inconsistent annotations for the same video~\cite{li2021temporal}.
%
}

\revised{To address the limitation of the full annotation, this paper introduces the \textbf{\textit{timestamp supervision}} to surgical phase recognition which trains the model from the timestamp annotation as shown in Fig.~\ref{fig:timestamp}~(a).
In timestamp annotation, the surgeons only annotate the phase class and a single timestamp for each phase, instead of start and end times. 
%
Once identifying the phase, the surgeon records the current timestamp (e.g., 00:05:20.00), no need to roll back and repeat watching the video to find precise start time.
After recording this single timestamp, since there is no need to find the end time, the surgeon would continue to go through the video quickly to find another phase.
Hence, the timestamp annotation significantly reduces manual annotation cost compared to the full annotations; see the detailed annotation analysis in Section~\ref{sec:annotation_analysis}.
%
%
}
%
%
%
\revised{Given timestamp supervision,~\emph{i.e.}, only a single label for each phase, the total number of positive frames is quite small, and the naive way that training with annotated labels may be difficult to learn a robust model; see results in Table.~\ref{tab:baseline}.}
%
%
%
{To generate more pseudo labels, some researchers propose to detect the action changes between two consecutive labeled frames for action recognition in natural videos with timestamp supervision~\cite{li2021temporal}.}
%
However, this method displays limited performance to surgical videos because surgical videos contain more ambiguous boundaries, leading to the noisy and inconsistent pseudo labels; see Sec.~\ref{sec:different_timestamp} for detailed discussion.

%
{To address the above problems, we leverage the the property of surgical videos to generate more trustworthy pseudo labels from timestamp supervision.}
%
{The property we observed is that phases in the surgical video are long events consisting of continuous frames, which shows a desirable temporal property that the closer the frames to the annotated timestamp, the more likely they are to be classified to the same label as the annotated one.}
Frames far from the annotated timestamp are difficult to have correct pseudo labels. 
%
%
Based on the above property, a \textbf{U}ncertainty-\textbf{A}ware \textbf{T}emporal \textbf{D}iffusion (\textbf{UATD}) module is proposed to diffuse the annotated timestamps to their adjacent low-uncertainty frames in the temporal axis.
In this way, only frames with high confidence and near the annotated timestamps would be considered for adding into pseudo-labels for training.
\revised{
Furthermore, the duration of the surgical videos generally lasts tens of minutes or even hours, making it hard to train the model in the end-to-end manner.
Current works~\cite{jin2017sv,jin2020multi,jin2021temporal} generally sample a few consequent frames from the long videos, and optimize the combined spatial-temporal model in the end-to-end manner.
This can be implemented in the full annotations, since all sampled consequent frames have labels.
However, in timestamp annotation, most of the sampled frames have no labels, resulting in the imbalance of positive and negative samples.
This imbalance training would degrade the performance; see details in Table~\ref{tab:ablative}.
To this end, we propose \textbf{L}oop \textbf{T}raining (\textbf{LP}), which optimizes the spatial and temporal model in an independent and iterative way.
%
}

{
We conduct empirical studies based on the proposed UATD and LP, and discover important insights of surgical phase recognition from timestamp supervision as follow:
\textbf{1)} Timestamp annotation can reduce $74\%$ annotation time compared with the full annotation, and surgeons tend to annotate those timestamps that are near the middle of phases; see details in Fig.~\ref{fig:annotation_analysis}.
\textbf{2)} Extensive experiments demonstrate that our method can achieve competitive results compared with full supervision methods, while reducing manual annotation cost; see details in Table~\ref{tab:sota}.
\textbf{3)} Less is more in surgical phase recognition,~\emph{i.e.}, less but discriminative pseudo labels outperform full but containing ambiguous frames; see details in -Table.~\ref{tab:sota}.
\textbf{4)} The proposed UATD can be used as a plug and play method to clean ambiguous labels near boundaries between phases, and improve the performance of the current surgical phase recognition methods; see details in Fig~\ref{fig:sim}. The reason is that training with our method would help to decrease intra-class distance and increase inter-class distance simultaneously; see details in Table.~\ref{tab:ours_full}.
}
The main contributions of this work can be summarized as the following:
\begin{itemize}
\item \revised{We study surgical phase recognition with a new time-stamp supervision, which is the the most efficient annotation setting in current surgical works.
We invite two surgeons with rich clinical experience to annotate timestamp annotations and record their annotation time, and find that the timestamp annotation can reduce $74\%$ annotation cost compared with the full annotation.
}
%
%
\item \revised{We introduce UATD to generate the trustworthy pseudo labels from the timestamp annotation, and LP to train the model from the generated pseudo labels in an iterative way.
}
\item \revised{ We conduct in-depth empirical studies of the proposed UATD and LP based on timestamp supervision, and discover four deep insights which may boost the future development of surgical phase recognition. Our code and timestamp annotations obtained from surgeons will be released at GitHub upon acceptance.}
%
%

\end{itemize}

\begin{figure*}[ht]
\centering
\includegraphics[width=0.85\textwidth,height=0.32\textheight]{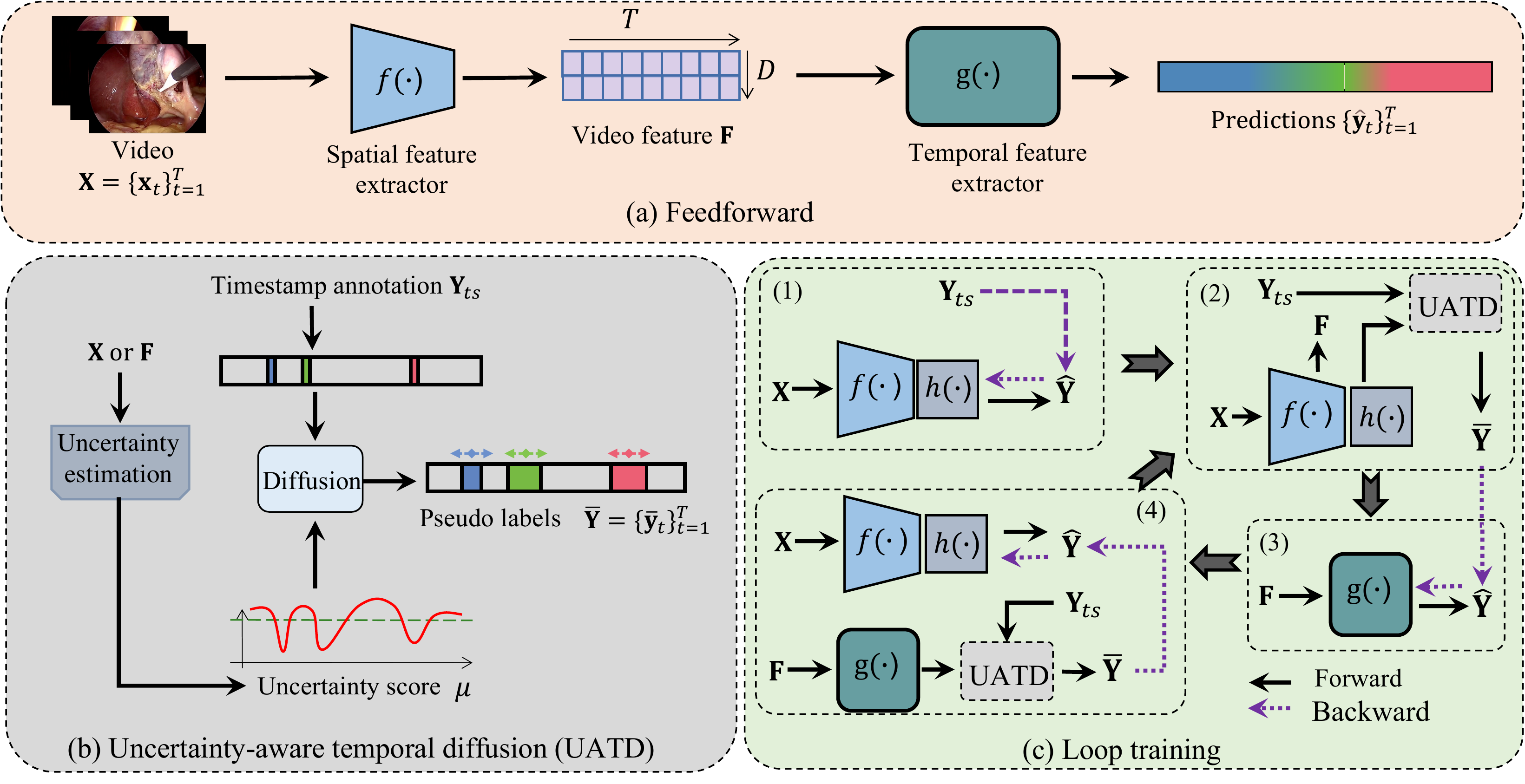}
\caption{
\revised{
Overview of our proposed framework.
(a) The feedforward process of mapping a video to the phase predictions. 
A video is first fed into a spatial feature extractor (normally a CNN) to obtain the video feature, followed by a temporal feature extractor to obtain the frame-wise prediction.
(b) Uncertainty-aware temporal diffusion (UATD). To generate trustworthy pseudo labels based on the timestamp supervision, videos or video features is fed into the uncertainty estimation to obtain the uncertainty scores for each frame.
Based on the uncertainty scores, we diffuse the timestamp annotation to the new pseudo labels.
(c) Loop training. Due to the computation and memory cost, the loop training is introduced to optimize the spatial feature extractor and temporal feature extractor by generated pseudo labels in an iterative way.  
}
}
\label{fig:our_method}
\end{figure*}

\section{Related Work}
\subsection{Surgical Phase Recognition}

We broadly classified related methods for surgical phase recognition into two categories including fully-supervised learning and label-efficient learning. 

\noindent \textbf{Fully-supervised Learning.} 
In fully-supervised learning, each frame in a surgical video is labeled.
Early works~\cite{blum2010modeling,lalys2011framework,graves2011practical} use hand-crafted features such as color and texture to perform recognition, which achieves limited performance and poor generalization.
With the development of neural networks, recent deep learning based methods achieve the great success~\cite{twinanda2016endonet,jin2017sv,funke2019using,yi2019hard,jin2020multi,jin2021temporal,yi2021not,zhang2021swnet,gao2021trans,ding2022exploring}.
\revised{
ZIBNET~\cite{sahu2020surgical} a state-preserving Long Short Term Memory (LSTM) to utilize the long-term evolution of tool usage within complete surgical phases.
}
EndoNet~\cite{twinanda2016endonet} first uses a convolutional neural network to automatically learn features and prove its effectiveness for surgical phase recognition. 
SV-RCNet~\cite{jin2017sv} integrates \revised{convolutional neural networks (CNN)} and \revised{long short-term memory (LSTM)} to learn both spatial and temporal representations in an end-to-end way.
To capture the long-range temporal relationship,
TMRNet~\cite{jin2021temporal} introduces a memory bank and TeCNO~\cite{czempiel2020tecno} uses dilated temporal convolutional network to get a large receptive field.
%
%
Recently, Yi~\emph{et al.}~\cite{yi2019hard} realize the negative effect of hard frames and propose data cleansing and online hard frames mapper to detect and handle them respectively.
Yi~\emph{et al.}~\cite{yi2021not} find that simply applying multi-stage architecture~\emph{e.g.} multi-stage TCN makes the refinement fall short and thus design not end-to-end training manner to alleviate this problem.
\revised{OperA~\cite{czempiel2021opera} leverages attention weight to yield further insights into the decision-making process.}
Trans-SVNet~\cite{gao2021trans} proposes a hybrid embedding aggregation Transformer to fuse spatial and temporal embedding.
Ding~\emph{et al.}~\cite{ding2022exploring} emphasize the importance of segment-level semantics and extract semantic-consistent segments to refine the erroneous predictions.
Notably, some related methods~\cite{twinanda2016endonet,czempiel2020tecno,jin2020multi} utilize additional tool presence labels to perform a multi-task learning to facilitate surgical phase recognition.

\noindent \textbf{{Semi-supervised Learning}.} Despite the great success the above methods get, they require a large amount of annotated videos, which is very costly.
{some researchers~\cite{yu2018learning,yengera2018less,dipietro2019automated,shi2020lrtd,shi2021semi} explore the methods for semi-supervision, where only parts of videos in the dataset are fully annotated, and others are unlabelled.
For example, LRTD~\cite{shi2020lrtd} use active learning to this context. It captures the long-range temporal dependency among continuous frames in the unlabeled data pool and selects the clips with weak dependencies to annotate.
Yengera~\emph{et al.}~\cite{yengera2018less} introduce self-supervised pre-training ensuring all available laparoscopic videos can be utilized.
Yu~\emph{et al.}~\cite{yu2018learning} propose a teacher/student approach where the teacher is trained on a small set of labeled videos and generates pseudo labels on the rest of unlabeled videos for student model learning.
Furthermore, SurgSSL~\cite{shi2021semi} uses consistency regularization and pseudo labeling to leverage the knowledge in unlabeled data, which progressively leverages the inherent knowledge held in the unlabeled data to a larger extent.
}
%
%
%
%
%

\noindent \textbf{\revised{Comparison of the manual annotation cost for different supervision setting}.}
\revised{
Here, we compare our proposed timestamp annotation compared with the above methods including full supervision methods and semi-supervision methods.
In full supervised methods~\cite{czempiel2020tecno,ding2022exploring}, annotators are required to repeat watching the video and roll back to find the precise start and end time for each phase, which is vert time-consuming.
As shown in Fig.~\ref{fig:timestamp}~(a), the average annotation time of each video for full supervision is $562.83$s.
In semi-supervision~\cite{yu2018learning,yengera2018less,dipietro2019automated,shi2020lrtd,shi2021semi}, the authors are required to only label full annotations for a few parts of all videos.
Generally, in semi-supervised surgical phase recognition methods, $50\%$ of videos are required to be annotated for achieving the competitive results compared with full supervision methods, as shown in Fig.~\ref{fig:timestamp}~(b).
However, the annotation times of the introduced timestamp supervision is only $148.53$s for each video,~\emph{i.e.}, $26\%$ annotation time of the full supervision, and achieve the competitive results.
For clarity, using the same network TCN~\cite{li2021temporal}, our methods achieve $91.9\%$ accuracy in Cholec80 with only $26\%$ annotation time, while the full supervision achieves $91.1\%$ accuracy using $100\%$ annotation time.
Meanwhile, the SOTA semi-supervised method SurgSSL~\cite{shi2021semi} achieves $87.0\%$ accuracy using $30\%$ annotation times.
Hence, our proposed method is the best trade-off between accuracy and manual annotation cost.
}


\if 
According to \cite{ma2020sf}, the annotation time for start and end frames,~\emph{i.e.}, full annotation is six times as large as that for timestamp labels.
\fi 
%

\subsection{{Weak} Supervision for Video Understanding}
Weakly supervision has received widespread attention in some video understanding tasks, such as temporal action localization~\cite{singh2017hide,wang2017untrimmednets,nguyen2018weakly,paul2018w,ding2020weakly,ding2021kfc,ding2021support} and action segmentation~\cite{bojanowski2014weakly,huang2016connectionist,li2019weakly}.
Some of them use video-level supervision, \emph{i.e.}, a set of action categories, while some use transcript-level supervision,~\emph{i.e.}, an ordered list of actions.
For example, Richard~\emph{et al.}~\cite{richard2018action} leverage text-based grammar from unordered action sets.
%
Although they significantly reduce the annotation effort, the performance is quite limited.
To trade-off the annotation-efficient and performance, timestamp supervision~\cite{mettes2016spot,moltisanti2019action,ma2020sf,li2021temporal} is proposed for action recognition.
%
%
For example, SF-Net~\cite{ma2020sf} designs an action frame mining and a background frame mining strategy to introduce more negative frames into the training process.
However, the above methods aiming at temporal action localization task generate very limited pseudo labels, is not suitable for surgical phase recognition,~\emph{i.e.}, frame-wise recognition.
In the action segmentation task, to generate frame-wise pseudo labels,  Li~\emph{et al.}~\cite{li2021temporal} detect the action change between two consecutive timestamps by stamp-to-stamp energy function and generate full pseudo labels.
%
However, in surgical videos, the frames near boundaries are generally ambiguous, the generated pseudo labels may be noisy annotations, which degrades the performance.
Compared with previous approaches, our proposed method generates as many confident pseudo labels as possible by considering the temporal relationships among frames, while discarding pseudo labels with a large uncertainty. 

\subsection{Uncertainty Estimation}
\revised{
In deep learning, neural networks may generate false prediction with a high probability, which is called epistemic uncertainty resulting from the model itself~\cite{kendall2017uncertainties}.
To estimate the uncertainty of the deep networks, Monte Carlo Dropout~\cite{gal2016dropout} is proposed to approximate the posterior distribution
for uncertainty estimation.
Ensembles~\cite{lakshminarayanan2017simple} trains multiple networks independently on the entire dataset using random, and the predictions of multiple networks are averaged over an ensemble.
Follow-up researchers majorly focus on improving the quality of the predicted uncertainty
scores by inference-based methods~\cite{tanno2017bayesian,jungo2020analyzing} or auto-encoder based methods~\cite{kingma2013auto,sohn2015learning}.
%
%
Estimation of uncertainty has also been investigated for medical image classification and
segmentation.
Laves~\emph{et al.}~\cite{laves2019uncertainty} leverages Monte Carlo dropout at test time, and shows that error prediction is correlated with higher uncertainty in OCT classification.
Leibig~\cite{leibig2017leveraging} uses Monte Carlo dropout to conduct uncertainty estimation and shows that uncertainty informed decision can improve the diagnostic performance. 
Wang~\emph{et al}~\cite{wang2019aleatoric} utilize Monte Carlo dropout and test
data augmentation to reduce overconfident error predictions in 3D brain tumor and 2D brain segmentation.
%
%
Different from current methods~\cite{leibig2017leveraging,wang2019aleatoric} that directly use Monte Carlo Dropout to estimate each sample individually, in our proposed UATD, the uncertainty of each frame is estimated based on the relation of itself, its nearby timestamp annotations and its adjacent frames in the temporal axis, which is motivated by the property of the surgical phase.
}

\section{Method}

\subsection{Problem Definition}~\label{sec:definition}
{
Let $\mathbf{X}=\{\mathbf{x}_t\}_{t=1}^T$ be a surgical video with $T$ frames, where \revised{$\mathbf{x}_t$} is the $t$-th frame.
Each surgical video is divided into several phases, and there is no overlapping among phases.
Our goal is to learn a spatial feature extractor network $f(\cdot)$ and a temporal feature extractor $g(\cdot)$ that maps the frame $\mathbf{x}_t$ to a phase label, which is presented in Fig.~\ref{fig:our_method}~(a).
In the full supervision, the frame-wise labels $\mathbf{Y} = \{ \mathbf{y}_1, \ldots, \mathbf{y}_T \}$ are available.
However, in our timestamp supervision, given a video consisting of $N$ phases, where $N << T$, only a single timestamp in each phase are annotated as $\mathbf{Y}_{ts} = \{ \mathbf{y}_{t_1}, \ldots, \mathbf{y}_{t_N} \}$, where $t_i$ is in the $i$-th phase, $\mathbf{y}_{t_i} \in \{ 1,2, \ldots, C \}$, and $C$ is the total number of classes.

}


{To perform surgical phase recognition with timestamp supervision, we propose an uncertainty-aware temporal diffusion (UATD) to generate trustworthy pseudo labels, denoted as $\overline{\mathbf{Y}}$, from the timestamp supervision $\mathbf{Y}_{ts}$ to optimize $f(\cdot)$ and $g(\cdot)$. The proposed UATD is shown in Fig.~\ref{fig:our_method}~(b); see Sec.~\ref{sec:uncertainty} for details.
%
%
Furthermore,  we introduce the loop training, which optimizes $f(\cdot)$ and $g(\cdot)$ in an iterative way to reduce the memory cost and imbalance optimization; see Fig.~\ref{fig:our_method}~(c) and Section~\ref{sec:training} for details.
}

\subsection{Uncertainty-aware Temporal Diffusion} \label{sec:uncertainty}
{
In timestamp supervision $\mathbf{Y}_{ts}$,~\emph{i.e.}, only a single label for each phase, the total number of positive frames is quite small and may be difficult to learn a robust model.
Although we do not have full annotations, it is clear that the phases are long events consisting of consecutive frames.
Motivated by this property of surgical videos, we propose the uncertainty-aware temporal diffusion (UATD) to diffuse the single labelled frame to its corresponding high confident (\emph{i.e.}, low uncertainty) neighbour frames.
In this way, we can introduce more frames acted as pseudo labels into the training process.
Furthermore, the diffusion of frames is stopped by low confident frames, which can avoid the ambiguous annotations.
The proposed UATD consists two components: uncertainty estimation and temporal diffusion.
In the following, we describe the two components respectively.
}
%

\revised{
\noindent\textbf{Uncertainty estimation.}
In UATD, we first need to estimated the uncertainty of each frame to find the high confident ones for the single annotated frame.
To this end, we introduce Monte Carlo Dropout~\cite{gal2016dropout}, a simple yet efficient way, to evaluate the uncertainty of each frame.
In Monte Carlo Dropout, given a input denoted as $\mathbf{z}$ and a network denoted as $o(\cdot)$, we feed $z$ into $o(\cdot)$ with different dropout $K$ times and obtain a set of class probabilities.
This process can be formulated as:
\begin{equation}
  \mathbf{P} = \{ \mathbf{p}^k=o(\mathbf{z}) \}_{k=1}^K,
  \label{E:ktimes}
\end{equation}
where $\mathbf{p}^k\in \mathbbm{R}^{C}$and $\mathbf{P}\in \mathbbm{R}^{K\times C}$, $C$ is the total number classes.
%
Then, we average these $K$ vectors of probability, which can be formulated as $\mu(\mathbf{P}) \in \mathbbm{R}^C$, where $\mu(\cdot)$ is the mean function.
After that, we obtain the class label for the input by:
\begin{equation}
c  ={\rm argmax}\;\mu(\mathbf{P}).
\label{E:class}
\end{equation}
Finally, we use the standard deviation to measure the uncertainty of the obtain class label,~\emph{i.e.}, $c$, which can be formulated as:
\begin{equation}
u  =\sigma\;(\mathbf{P}_{c}),
\label{E:uncertainty}
\end{equation}
where $u$ is the uncertainty score for $o(\cdot)$ with the input of $\mathbf{z}$.
The higher $u$ indicates that the model $o(\cdot)$ predicts $\mathbf{z}$ to class $c$ with lower confidence, and vice versa.
In this paper, we need evaluate the uncertainty of both the spatial and temporal feature extractors, which are defined in Section~\ref{sec:definition}.

To conduct the uncertainty estimation for the spatial feature extractor $f(\cdot)$, we add an extra classification head $h(\cdot)$ to $f(\cdot)$ as shown in Fig.~\ref{fig:our_method}~(c), denoted as $h(f(\cdot))$ to obtain the classification prediction for each frame $\mathbf{x}_t$.
Let $o(\cdot) = h(f(\cdot))$ and $\mathbf{z} = \mathbf{x}_t$, and then we can obtain the uncertainty score $\mu_t$ for each frame $\mathbf{x}_t$ by using Eq.~\ref{E:ktimes} to Eq.~\ref{E:uncertainty}.
Similarly, to conduct the uncertainty estimation for the spatial feature extractor $g(\cdot)$, we can easily set $o(\cdot) = g(\cdot)$ and $\mathbf{z} = \mathbf{f}_t$, obtaining the uncertainty score for each frame feature $\mathbf{f}_t$.
}

	
		
			
			
	
	
\revised{
\begin{algorithm}[t]
\caption{Temporal Diffusion}
\label{alg:diffsuion}
\textbf{Input:}%
Uncertainty scores $\{\mathbf{\mu}_t\}_{t=1}^T$,
prediction $\hat{\mathbf{Y}} = \{{\hat{\mathbf{y}}}_t \}_{t=1}^T$,
timestamp annotation $\mathbf{Y}_{ts} = \{ \mathbf{y}_{t_1}, \ldots, \mathbf{y}_{t_N} \}$, uncertainty threshold $\tau$. \\
%
\textbf{Output:}%
Pseudo labels $\overline{\mathbf{Y}} = \{\overline{\mathbf{y}}_t \}_{t=1}^T$ for the next iteration.
%
\begin{algorithmic}[1]
\LineComment{Diffusion for each phase}
\For{$i=1$ to $N$} 
    \LineComment{Diffusion for the left side}
    \For{$t= t_{i-1}$ to $t_i$}
        \State $\overline{\mathbf{y}}_{t} = \hat{\mathbf{y}}_t \cdot \mathbbm{1}(u_t < \tau) \cdot \mathbbm{1}(\hat{\mathbf{y}}_t = \mathbf{y}_{t_i})$
    \EndFor
    \LineComment{Diffusion for the right side}
    \For{$t= t_{i}$ to $t_{i+1}$}
        \State $\overline{\mathbf{y}}_{t} = \hat{\mathbf{y}}_t \cdot \mathbbm{1}(u_t < \tau) \cdot \mathbbm{1}(\hat{\mathbf{y}}_t = \mathbf{y}_{t_i})$
    \EndFor
\EndFor
\end{algorithmic}
\end{algorithm}
}

{
\noindent{\bf Temporal diffusion.} After obtaining the uncertainty score $\mu_t$, we use the temporal diffusion module to diffuse the current labels to more pseudo labels for training in the next iteration; see the iterative training details in Sec.~\ref{sec:training}.
%
%
To be specific, we treat the labeled frames as anchors and start diffusion from anchors to the adjacent frames on either sides of them in temporal dimension, which is illustrated in Algorithm~\ref{alg:diffsuion}.
%
%
By the temporal diffusion, one frame would be introduced into next iteration training only if the uncertainty score of it is lower than a threshold~$\tau$ and the predicted class label equals to its nearby timestamp frame.
In this way, the generated pseudo label would be high confidence, avoid introducing noisy annotations.
Note that in the obtained pseudo labels $\overline{\mathbf{Y}} = \{\overline{\mathbf{y}}_t \}_{t=1}^T$, $\mathbf{y}_t = 0$ means the $t$-th frame is not labelled.
}

\revised{
For clarity, we formulate the overall process of UATD as $\overline{\mathbf{Y}} \gets \text{UATD}(o(\cdot), \mathbf{Z}, \hat{\mathbf{Y}}, \tau, K)$, where $\mathbf{Z}$ is the input (~\emph{e.g.}, $\mathbf{X}$ or $\mathbf{F}$), $o(\cdot)$ is the network (~\emph{e.g.}, $h(f(\cdot))$ or $g(\cdot)$).
}
\subsection{Loop Training}\label{sec:training}

\begin{algorithm}[t]
\caption{\revised{Loop training}}
\label{alg:training}
\textbf{Input:} \revised{Video $\mathbf{X}$,  timestamp annotation $\mathbf{Y}_{ts}$, initial spatial feature extractor $f(\cdot)$, initial spatial classifier $h(\cdot)$, initial temporal feature extractor $g(\cdot)$, uncertainty threshold $\tau$, forward times $K$, times of temporal diffusion $n$, times of loop training $m$}. \\
\textbf{Output:} \revised{Well optimized spatial feature extractor $f(\cdot)$ and temporal feature extractor $g(\cdot)$.}
\begin{algorithmic}[1]

\State \revised{$\overline{\mathbf{Y}} \gets \mathbf{Y}_{ts}$ \Comment{Set the initial pseudo labels}}
\For{$i=1$ to $n$} 
    \LineComment{\revised{Optimizing the spatial feature extractor}}
    \State \revised{$f(\cdot),h(\cdot) \gets \text{OptimS}(f(\cdot),h(\cdot),\mathbf{X},\overline{\mathbf{Y}}, \mathcal{L}_{ce})$}
    \LineComment{\revised{Use UATD to generate the new pseudo labels}}
    \State  \revised{$\overline{\mathbf{Y}} \gets \text{UATD}(h(f(\cdot)), \mathbf{X}, {\mathbf{Y}_{ts}}, \tau, K)$.
    }
\EndFor
\For{$j=1$ to $m$} 
\State \revised{$\mathbf{F}\gets f(\mathbf{X})$}
\For{$i=1$ to $n$} 
    \LineComment{\revised{Optimizing the spatial feature extractor}}
    \State \revised{$g(\cdot) \gets \text{OptimT}( g(\cdot),\mathbf{F},\overline{\mathbf{Y}}, \mathcal{L}_{ce}, \mathcal{L}_{smooth})$}
     \LineComment{\revised{Use UATD to generate the new pseudo labels}}
    \revised{\State $\overline{\mathbf{Y}} \gets$ UATD($g(\cdot)$, $\mathbf{F}$, $\mathbf{Y}_{ts}$, $\tau$, $K$)}
\EndFor
\LineComment{\revised{Optimizing the spatial feature extractor}}
\State \revised{$f(\cdot),h(\cdot) \gets \text{OptimS}(f(\cdot),h(\cdot),\mathbf{X},\overline{\mathbf{Y}}, \mathcal{L}_{ce})$}
\EndFor
\end{algorithmic}
\end{algorithm}

\revised{The duration of the surgical videos generally lasts tens of minutes or even hours, making it hard to train the model in the end-to-end manner.
In previous full supervised methods~\cite{jin2017sv,jin2020multi,jin2021temporal}, a few consequent frames are sampled from the long videos for training the spatial and temporal networks in the end-to-end manner.
However, in timestamp annotation, most of the sampled frames have no labels, resulting in the imbalance of positive and negative samples.
This imbalance training would degrade the performance; see details in Table~\ref{tab:ablative}.
To address this problem, we decouple the optimization of spatial and temporal feature extractors via loop training, as shown in Fig.~\ref{fig:our_method}~(c).}
In the loop training, we only sample labelled frames (annotated timestamp or generated pseudo labels) to optimize the spatial feature extractor or temporal feature extractor, which can not be achieved in previous jointly training.
%
{
Formally, there four main steps in our loop training: 

\textbf{(a)} Optimizing the spatial feature extractor: $f(\cdot),h(\cdot) \gets \text{OptimS}(f(\cdot),h(\cdot),\mathbf{X},\overline{\mathbf{Y}}, \mathcal{L}_{ce})$.
To be specific, the input video $\mathbf{X}$ is fed into the spatial feature extractor $f$ to obtain the video feature $\mathbf{F} = f(\mathbf{X})$. Then a classifier $h(\cdot)$ is used to obtain the prediction $\hat{\mathbf{Y}} = h(\mathbf{F})$, where $\hat{\mathbf{Y}} = \{ \hat{\mathbf{y}_t} \}_{t=1}^T$.
Given the target labels (timestamp annotation or pseudo labels) $\overline{\mathbf{Y}} = \{ \overline{\mathbf{y}}_t \}_{t_1}^T$, the objective for the spatial feature extractor can be formulated as:
\begin{equation}
 \mathcal{L}_{ce} =  -\frac{1}{T} \sum_{t=1,\overline{\mathbf{y}}_{t} \neq0 }^{T}  \overline{\mathbf{y}}_{t} \log (\hat{\mathbf{y}}_{t}),
     \label{ce1}
\end{equation}
where $\overline{\mathbf{y}}_{t} \neq0$ indicates the $t$-th frame is not labelled.

\textbf{(b)} Extracting the spatial features: $\mathbf{F} = f(\mathbf{X})$; see details in step \textbf{(1)}.

\textbf{(c)} Optimizing the temporal feature extractor: $g(\cdot) \gets \text{OptimT}( g(\cdot),\mathbf{F},\overline{\mathbf{Y}}, \mathcal{L}_{ce}, \mathcal{L}_{smooth})$.
Specifically, the video feature $\mathbf{F}$ is fed into $g(\cdot)$ to capture the temporal relation of frames and obtain the corresponding predictions $\hat{\mathbf{Y}}$.
We use the CrossEntropy loss to train the  $g(\cdot)$, similar as  $f(\cdot)$.
}
Compared with the spatial feature extractor, to encourage a smooth transition between frames, we use the truncated mean squared error as a {Smoothing Loss} following~\cite{farha2019ms,li2021temporal}:
\begin{align}
    \mathcal{L}_{smooth}=\frac{1}{TC}\sum_{t,c}\Tilde{\Delta}_{t,c}^2, \\
    \Tilde{\Delta}_{t,c}^2=\left\{
    \begin{aligned}
    \Delta_{t,c}^2&,&\quad\Delta_{t,c} < \gamma \\
    \gamma&,&\quad otherwise
    \end{aligned}
    \right., \\
    \Delta_{t,c}=|log({\hat{\bm{y}}_{t,c}})-log(\hat{\bm{y}}_{t-1,c})|,
    \label{smooth}
\end{align}
where $T$ is the video length and $C$ is the number of action classes. This loss function explicitly penalize the difference of each two adjacent frames and we suggest readers refer to~\cite{farha2019ms} for more details.
The final loss function is the weighted sum of these two losses:
\begin{equation}
    \mathcal{L}=\mathcal{L}_{ce} + \lambda \mathcal{L}_{smooth},
\end{equation}
where $\lambda$ is a hyper-parameter to balance the contribution of each loss and is set to $0.015$ for all of our experiments.

\revised{
\textbf{(d)} Generating the pseudo labels based on UATD: $\overline{\mathbf{Y}} \gets \text{UATD}(o(\cdot), \mathbf{Z}, {\mathbf{Y}}_{ts}, \tau, K)$; see details in Section~\ref{sec:uncertainty}.
}

\noindent{After the definiation of the four steps, we illustrate the loop training in Algorithm~\ref{alg:training}.
%
}

\begin{figure}[t]
\centering
\includegraphics[width=0.5\textwidth,height=0.25\textheight]{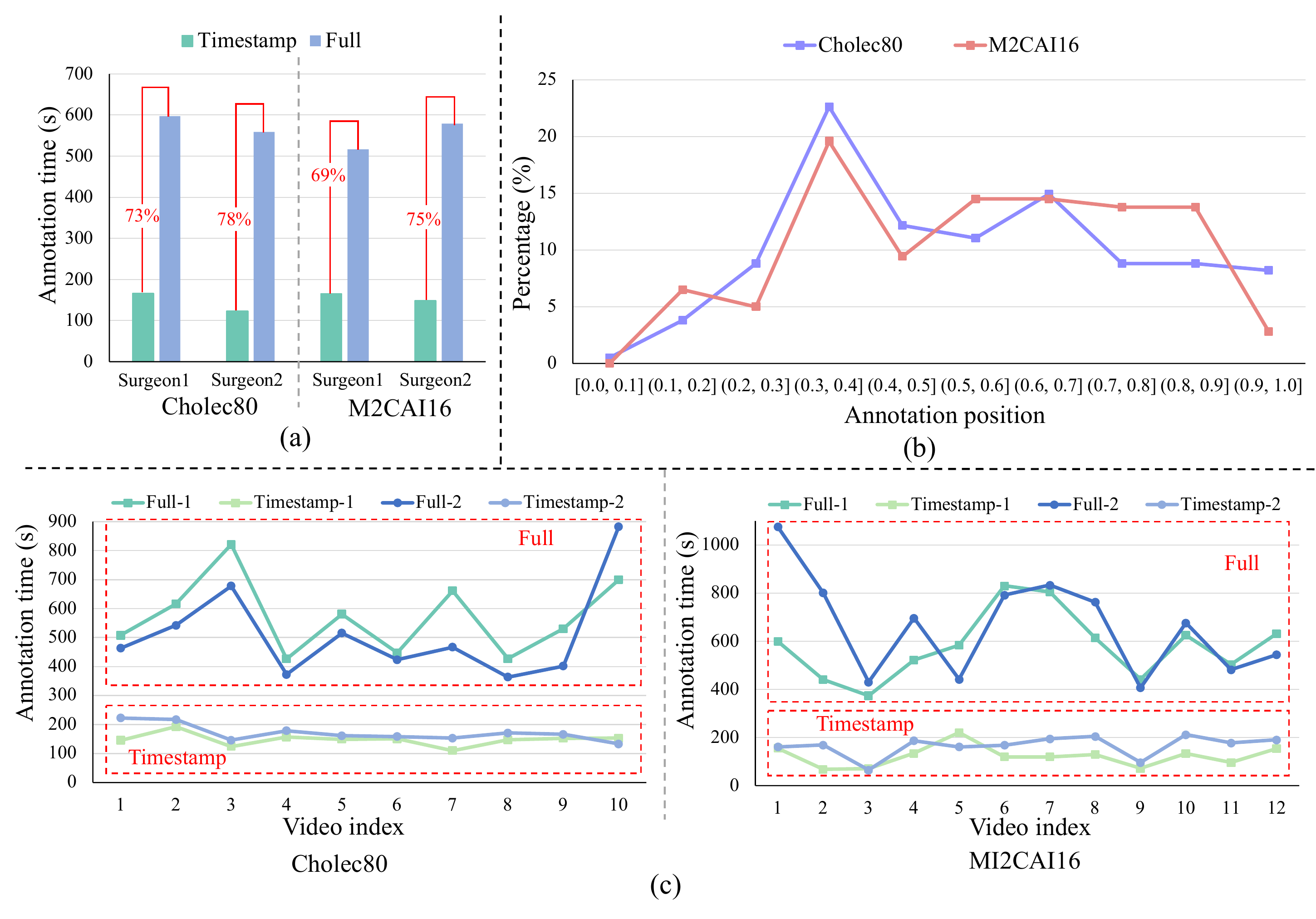}
\caption{
\revised{(a) Comparison of the annotation time of timestamp and full annotations.
We show the annotation times of two different surgeons for Cholec80 and M2CAI16 datasets, which are reported as seconds/video for each dataset.
(b) Statistics of positions of manually annotated timestamps on two datasets.
The horizontal axis indicates the relative temporal portion of the whole phase.
For example, (0.1, 0.2] indicates the annotated timestamp is inside the temporal period from 0.1 to 0.2 of the phase. The vertical axis represents the percentage of annotated timestamps.
It shows that the timestamps would appear in the arbitrary position of the phase, and surgeons prefer to label timestamp near the middle of phases.
(c) Statistics of annotation time of manually annotated timestamps on two datasets.
The horizontal axis indicates the video index, and the vertical axis represents annotation time.
It shows that our timestamp annotation consistently takes less time than full annotation for labeling each video.
}
}
\label{fig:annotation_analysis}
\end{figure}
\begin{table*}[ht]\normalsize
\centering
\caption{Comparison with the state-of-the-art on Cholec80 and M2CAI16 datasets. \revised{``$^*$" indicates the offline prediction.}}
\label{tab:sota}
\resizebox{2.0\columnwidth}{!}{
\begin{tabular}{l|rrrr|rrrr}
\hline
\multirow{2}{*}{Method} & \multicolumn{4}{c}{Cholec80} & \multicolumn{4}{c}{M2CAI16} \\
 & AC (\%) & PR (\%) & RE (\%) & JA (\%) & AC (\%) & PR (\%) & RE (\%) & JA (\%) \\
\hline
\multicolumn{9}{c}{Fully Supervised Methods - $\bm{100\%}$ annotation time}\\ 
\hline
PhaseNet~\cite{twinanda2016single} & $78.8\pm4.7$ & $71.3\pm15.6$ & $76.6\pm16.6$ & - & $79.5\pm12.1$ & - & - & $64.1\pm10.3$ \\
EndoNet~\cite{twinanda2016endonet} & $81.7\pm4.2$ & - & $79.6\pm7.9$ & - & - & - & - & - \\
SV-RCNet~\cite{jin2017sv} & $85.3\pm7.3$ & $80.7\pm7.0$ & $83.5\pm7.5$ & - & $81.7\pm8.1$ & $81.0\pm8.3$ & $81.6\pm7.2$ & $65.4\pm8.9$ \\
OHFM\cite{yi2019hard} & $87.3\pm5.7$ & - & - & $67.0\pm13.3$ & $85.2\pm7.5$ & - & - & $68.8\pm10.5$ \\
 Casual TCN~\cite{li2021temporal} & $87.9\pm8.2$ & $86.4\pm7.7$ & $84.8\pm{12.9}$ & $72.4\pm{9.4}$ & $81.9\pm{11.3}$ & $84.8\pm{5.2}$ & $82.2\pm{9.0}$ & $68.1\pm{8.5}$ \\
TeCNO~\cite{czempiel2020tecno} & $88.6\pm7.8$ & $86.5\pm7.0$ & $88.8\pm17.4$ & $75.1\pm6.9$ & - & - & - & - \\
TMRNet~\cite{jin2021temporal} & $90.1\pm7.6$ & $90.3\pm3.3$ & $89.5\pm5.0$ & $79.1\pm5.7$ & $87.0\pm8.6$ & $87.8\pm6.9$ & $88.4\pm5.3$ & $75.1\pm6.9$ \\
Trans-SVNet~\cite{gao2021trans} & $90.3\pm7.1$ & $90.7\pm5.0$ & $88.8\pm7.4$ & $79.3\pm6.6$ & $87.2\pm9.3$ & $88.0\pm6.7$ & $87.5\pm5.5$ & $74.7\pm7.7$ \\
%
TCN\revised{$^*$}~\revised{\cite{farha2019ms}} & $91.1\pm{6.7}$ & ${90.8\pm{4.5}}$ & $87.6\pm{11.7}$ & $79.1\pm{8.5}$ & $82.9\pm{10.8}$ & $85.8\pm{5.4}$ & $82.7\pm{9.0}$ & $69.7\pm{8.7}$ \\
Not end-to-end~\cite{yi2021not} & $91.5\pm7.1$ & - & $87.2\pm8.2$ & $77.2\pm11.2$ & ${88.2\pm8.5}$ & - & ${91.4\pm11.2}$ & $75.1\pm10.6$ \\

\hline
\multicolumn{9}{c}{Semi Supervised Methods - $\bm{50\%}$ annotation time}\\ 
\hline
LRTD~\cite{shi2020lrtd} & $82.5\pm8.4$ & $79.7\pm9.0$ & $80.9\pm8.1$ & $64.2\pm10.2$ & $72.1\pm13.7$ & $74.1\pm14.9$ & $74.0\pm10.4$ & $54.4\pm12.9$ \\
SurgSSL~\cite{shi2021semi} & $87.0\pm7.4$ & $84.2\pm8.9$ & $85.2\pm11.1$ & $70.5\pm12.6$ & $79.6\pm9.4$ & $80.2\pm11.3$ & $79.6\pm11.5$ & $62.0\pm11.1$ \\
\hline
\multicolumn{9}{c}{Timestamp Supervised Methods - \revised{$\bm{26\%}$} annotation time}\\ 
\hline
 Casual TCN+Ours & $88.6\pm6.7$ & $86.1\pm6.7$ & $88.0\pm10.1$ & $73.7\pm10.2$ & $86.0\pm7.8$ & $85.0\pm6.2$ & $87.1\pm7.7$ & $71.4\pm10.4$ \\
TCN\revised{$^*$}+Ours & ${91.9\pm5.6}$ & $89.5\pm4.4$ & ${90.5\pm5.9}$ & ${79.9\pm8.5}$ & $87.6\pm8.7$ & ${88.2\pm7.4}$ & $87.9\pm9.6$ & ${75.7\pm9.5}$ \\
\hline
\end{tabular}}
\end{table*}
\section{Experiments}

\subsection{Datasets and Metrics}

\noindent \textbf{M2CAI16.} The M2CAI16 dataset~\cite{twinanda2016miccai} consists of $41$ laparoscopic videos that are acquired at 25fps of cholecystectomy procedures, and each frame has a resolution of 1920$\times$1080. Following ~\cite{yi2021not}, $27$ videos are used for training while the rest $14$ are used for testing. These videos are segmented into $8$ phases by experienced surgeons.

\noindent \textbf{Cholec80.} The cholec80 dataset~\cite{twinanda2016endonet} contains $80$ videos of cholecystectomy surgeries performed by $13$ surgeons. All the videos are recorded at $25$ fps, and the frames in them have the resolution of $1920\times1080$ or $854\times480$. The dataset is divided into two subsets of equal size, with the first $40$ videos as a training set and the other $40$ as a testing set.

\noindent \textbf{Evaluation metrics.} Following previous works~\cite{jin2017sv,jin2021temporal,twinanda2016endonet,yi2019hard}, we utilize four metrics,~\emph{i.e.}, accuracy (AC), precision (PR), recall (RE), and Jaccard (JA), to evaluate the phase prediction accuracy. Among them, accuracy and Jaccard index are used to evaluate the submission of M2CAI Workflow Challenge, while precision and recall are also commonly used metrics for video-based phase recognition.

\subsection{\revised{Annotation Analysis}}~\label{sec:annotation_analysis}
\revised{
To obtain the timestamp annotations, we invite two surgeons to label a single timestamp for each phase on two datasets.
Specifically, they are asked to label one timestamp for each phase while watching the video, as shown in Fig.~\ref{fig:timestamp}~(a).
To compare the annotation cost of different types of annotations, we also ask them to find the precise the start and end time for each phase,~\emph{i.e.}, full annotation.
In Fig.~\ref{fig:annotation_analysis}~(a), we report the average time they spend on each video when using timestamp and full annotations.
``Surgeon1" and ``Surgeon2" indicates the first surgeon and the second surgeon respectively.
To obtain annotation times for full or timestamp annotations, we first let the surgeon prepare a timer. When conducting full or timestamp annotations, the surgeon first turned on the timer, then immediately watched the video and annotated it. After completing the annotation of a video, the surgeon stopped the timer immediately, and record the time it takes to annotate the video. When all videos are annotated and their annotation time are recorded, we calculate the average annotation time for all videos.
It is clear that our introduced timestamp annotation can largely reduce the annotation time compared with the full annotation,~\emph{e.g.},Surgeon2 can reduce $78\%$ time in Cholec80 dataset.
On average, our proposed timestamp annotation only requires $26\%$ annotation time compared with the full annotation.

Furthermore, we also show the distribution of the relative position of timestamp annotation to the corresponding phase on two datasets.
As shown in Fig.~\ref{fig:annotation_analysis}~(b), the labeled timestamps would appear in arbitrary position of the phase.
Surgeons prefer to label timestamp near the middle of phases, which reveals that the surgeons can identify a phase without watching the whole phase.
That is to say, the surgeons can skip the left part of the phase after the timestamp annotation.
Of course, there is no need for the surgeons to repeat watching videos to find precise the temporal window for each phase.
Hence, the annotation time can largely be reduced compared with the full annotation.
In the implementation, one second of video is converted to $25$ frames.
To save memory and computation cost, we sample one frame every second. Hence, during annotation, the surgeon labels the second, and during implementation, we set the frame belonging to the timestamp second as the annotation.
Finally, we show the statistics of annotation time of manually annotated timestamps on two datasets in Fig.~\ref{fig:annotation_analysis}~(c).
The results show that the annotation times of timestamp are much less than those of full for all videos.

%
%
}

\subsection{Implementation Details}
Our code is based on PyTorch using an NVIDIA GeForce RTX 3090 GPU.
We downsample the video to 1fps for training in all experiments following previous works~\cite{twinanda2016endonet,jin2017sv,jin2021temporal}.
All the frames are resized to a resolution of $250\times 250$, and data augmentations including $224\times224$ cropping, random mirroring, and color jittering are applied during the training stage.
We get a pre-trained inception-v3~\cite{szegedy2016rethinking} on ImageNet~\cite{krizhevsky2012imagenet}.
The batch size is set to $8$, and an Adam optimizer with an initial learning rate of $1e-4$ and weight decay of $1e-5$ is used.
We further use a step learning rate scheduler where the step size is two epochs and decay rate is $0.5$ for fune-tuning by $5$ epochs.
To train TCN, we use Adam optimizer with an initial learning rate of $1e-3$ and cosine annealing for learning rate decay.
For all experiments, we set a dropout rate of $0.5$ and an uncertainty threshold $\tau=0.1$; the detailed analysis is shown in Table~\ref{tab:thres}.
The uncertainty is estimated by $5$ forward times Monte Carlo Dropout.~\cite{gal2016dropout}.
The numbers of rounds of uncertainty-aware temporal diffusion and loop training are set to $m=4$ and $n=2$, respectively.
Furthermore, the timestamp annotations are simulated by randomly selecting one frame from each action phase in the training videos.

\begin{figure*}[t]
\centering
\includegraphics[width=1\textwidth,height=0.32\textheight]{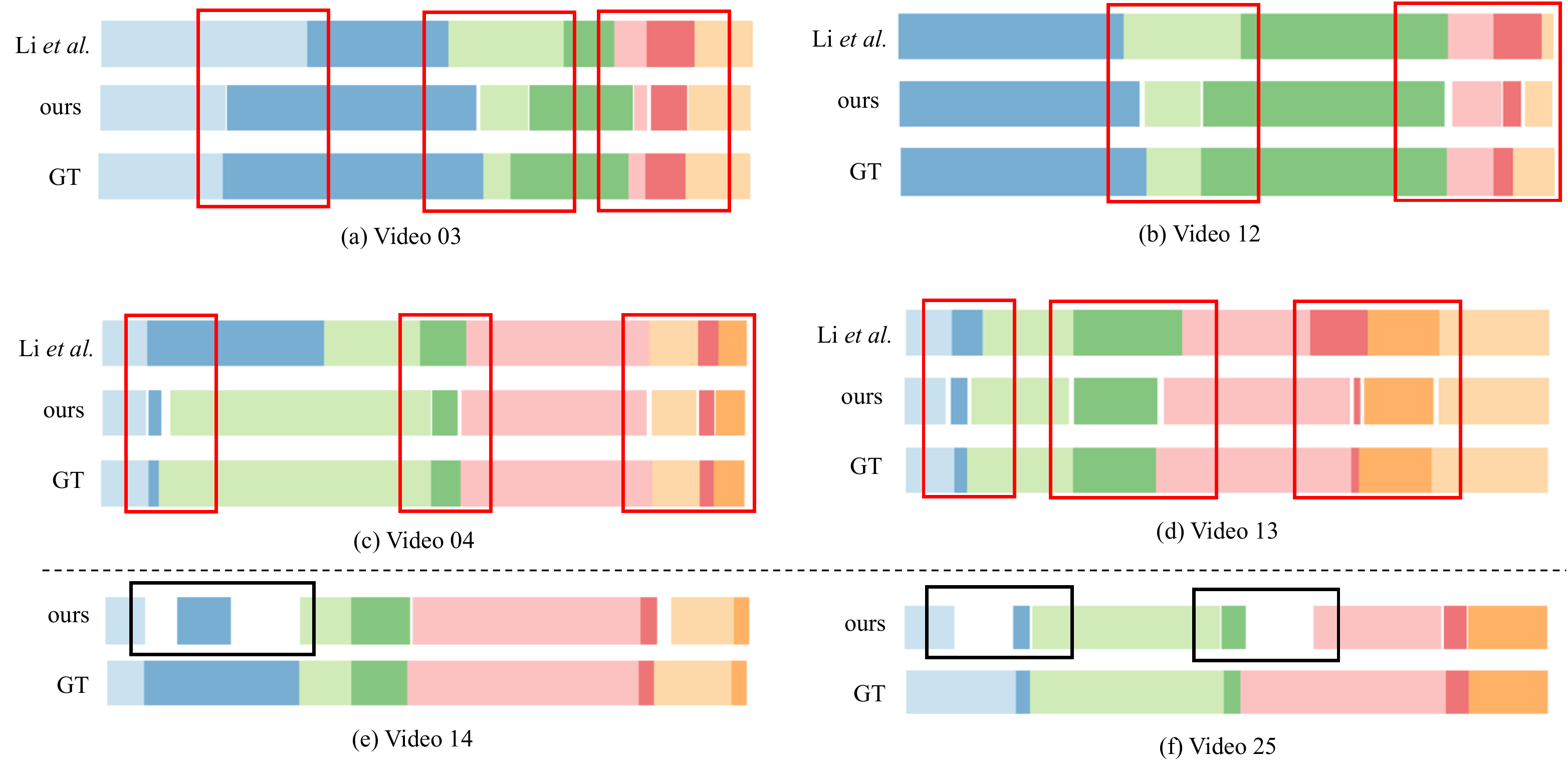}
\caption{ 
\revised{
Comparison of the visualization of pseudo labels generated by ours and Li~\emph{et al.}~\cite{li2021temporal}.
``GT" indicates the ground-truth.
We sample four videos,~\emph{i.e.}, from Cholec80 ((a)-(b)) and M2CAI16 ((c)-(d)). It is clear that our method can generate more accurate pseudo labels compared with Li~\emph{et al.}~\cite{li2021temporal}; see red boxes.
We also illustrate the worst pseudo labels generated by our method shown in (e)-(f). The uncertainty frames inside the phases would make the model to stop temporal diffusion (see black boxes).
}
}
\label{fig:vis}
\end{figure*}
\subsection{Comparison with the State-of-the-Arts}
We compare our \emph{less is more} method with the state-of-the-arts on the Cholec80 and M2CAI16 datasets, and report their results in Table~\ref{tab:sota}.
\revised{
The numbers in Table~\ref{tab:sota} are the mean and standard deviation of performance of all phases.
For example, in Cholec80, there are 7 phases. To obtain the accuracy (AC) for each method, we first obtain AC for each phase, the mean of AC of 7 phases is computed to obtain the first number in Table~\ref{tab:sota}. After that, we compute the standard deviation of AC numbers of 7 phases to obtain the number after ``+/-". The computation of PR, RE and JA is like AC. 
}
It is clear that our method outperforms previous data-efficient methods,~\emph{i.e.}, semi-supervised ones, on both data efficiency and phase recognition performance.
For example, our timestamp supervision only requires \revised{$26\%$} annotation time of the full supervision~\cite{ma2020sf}, while semi-supervision needs $50\%$ annotation time~\cite{shi2020lrtd}.
Moreover, our method with the casual TCN~\cite{li2021temporal} achieves $88.6\%$ of accuracy on Cholec80 dataset, achieving the \revised{competitive performance} compared to semi-supervised methods.
%
We can also find that our method can even achieve the competitive performance compared with the fully supervised methods, with only \revised{$26\%$} annotation time of them.
%
%
%
%
Notably, the improvements of our method are more significant in M2CAI16 than in Cholec80.
This is because M2CAI16 contains more ambiguous frames~\cite{ding2022exploring}, which degrades the performance.
The details why our methods can outperform corresponding backbones in fully supervised setup will be discussed in Sec.~\ref{sec:TvsF}.
\begin{table}[t]
\centering
\caption{Comparison with different timestamp supervision methods.}
\label{tab:baseline}
\resizebox{1.0\columnwidth}{!}{
\begin{tabular}{l | rrrr}
\hline
 Method & AC (\%) & PR (\%) & RE (\%) & JA (\%) \\
\hline
\multicolumn{5}{c}{Cholec80} \\
\hline
Naive &\revised{ $66.9\pm5.6$} & \revised{$62.3\pm6.5$} & \revised{$74.8\pm6.5$} & \revised{$48.2\pm4.4$} \\
Uniform & \revised{$58.7\pm 7.8$} & \revised{$55.5 \pm 6.1$} & \revised{$65.9 \pm 5.4$} & \revised{$39.0 \pm 6.5$} \\
Li~\emph{et al.}\protect\cite{li2021temporal} & \revised{$79.4 \pm 5.5$} & \revised{$78.7 \pm 6.5$} & \revised{$85.4 \pm 5.5$} & \revised{$64.0 \pm 5.6 $} \\
Ours  &  \revised{${88.6\pm6.7}$} &  \revised{$86.1\pm6.7$} & \revised{$88.0\pm10.1$} & \revised{$73.7\pm10.2$} \\
\hline
\multicolumn{5}{c}{M2CAI16} \\
\hline
Naive & \revised{$67.5 \pm 7.2$} & \revised{$58.7 \pm 6.5$} & \revised{$61.7 \pm 6.5$} & \revised{$ 44.8 \pm 6.7$} \\
Uniform & \revised{$56.5 \pm 8.7$} & \revised{$56.7 \pm 7.7$} & \revised{$57.0 \pm 7.9$} & \revised{$38.2 \pm 5.6$} \\
Li~\emph{et al.}\protect\cite{li2021temporal} & \revised{$72.7 \pm 8.8$} & \revised{$76.5 \pm 7.1$} & \revised{$80.5 \pm 6.9$} & \revised{$59.9 \pm 10.1$} \\
Ours  & \revised{$86.0\pm7.8$} & \revised{$85.0\pm6.2$} & \revised{$87.1\pm7.7$} & \revised{$71.4\pm10.4$} \\
\hline
\end{tabular}}
\end{table}
\subsection{Comparison with Different Timestamp Supervision Methods}
\label{sec:different_timestamp}

To evaluate the efficiency of our proposed uncertainty-aware temporal diffusion (UATD) for surgical video timestamp supervision, we compare our methods with two baseline models following~\cite{li2021temporal},~\emph{i.e.}, Naive and Uniform, and report the results in Table~\ref{tab:baseline}.
\revised{Specifically, in Naive, we only use the annotated timestamp labels to supervise the model training, without generating any pseudo labels.
In Uniform, the pseudo labels are generated by a uniform way,~\emph{i.e.}, the action labels change at the center frame between two timestamp annotations.
For example, assuming two timestamps $\mathbf{y}_{t_1}$ and $\mathbf{y}_{t_2}$ with $t_1 < t_2$, then the pseudo labels can be generated as:
\begin{equation}
\hat{\mathbf{y}}_{t} =\left\{
    \begin{aligned}
     \mathbf{y}_{t_1} &, & t \in (t_1, t_1 + (t_2 - t_1) / 2 ] \\
     \mathbf{y}_{t_2} &, & t \in (t_1 + (t_2 - t_1) / 2, t_2 )
    \end{aligned}
    \right..
\end{equation}
}
%
It is clear that our method outperforms other two methods with a clear margin.
Furthermore, we also compare Li~\emph{et al.}~\cite{li2021temporal}, which is the SOTA in action segmentation under this setting.
It uses the middle output of model~\emph{i.e.}, features of frames to detect action change and generate frame-wise pseudo labels
However, using feature similarity to detect action change could be confused when the boundaries are generally ambiguous.
As shown in Fig. \ref{fig:vis}, our methods give accurate pseudo labels by stopping diffusion near boundaries while \cite{li2021temporal} attempts to give unappealing labels.
From Table. \ref{tab:baseline}, we can also see that our method obtains $7\%-13\%$ improvements over all metrics.
%

\begin{table}[t]\normalsize
\centering
\caption{Ablative study of key components on Cholec80 dataset. `UATD (S)' and `UATD (T)' indicate using UATD in the spatial feature and temporal feature extractors. `LP' indicates the loop training which is defined in Sec.~\ref{sec:training}.}
\label{tab:ablative}
\resizebox{1.0\columnwidth}{!}{
\begin{tabular}{c c c|rrrr}
\hline
UATD (S) & UATD (T) & LP & AC (\%) & PR (\%) & RE (\%) & JA (\%) \\
\hline
{\XSolidBrush} & {\XSolidBrush} & {\XSolidBrush} & \revised{$66.9 \pm 9.7$} &  \revised{$62.3 \pm 6.7$} & \revised{$74.8 \pm 7.8$} & \revised{$48.2 \pm 7.6$} \\
{\CheckmarkBold} & {\XSolidBrush} & {\XSolidBrush} & \revised{$82.3 \pm 7.6$} & \revised{$78.1 \pm 8.8$} & \revised{$86.9 \pm 6.5$} & \revised{$66.0 \pm 7.4$} \\
{\XSolidBrush} & {\CheckmarkBold} & {\XSolidBrush} & \revised{$77.6 \pm 5.3$} & \revised{$77.3 \pm 6.7$} & \revised{$81.0 \pm 7.3$} & \revised{$61.3 \pm 5.2$} \\
{\XSolidBrush} & {\XSolidBrush} & {\CheckmarkBold} & \revised{$68.5 \pm 4.8$} & \revised{$63.7 \pm 6.2$} & \revised{$75.2 \pm 3.7$} & \revised{$50.2 \pm 6.1$} \\
{\CheckmarkBold} & {\CheckmarkBold} & {\XSolidBrush} & \revised{$85.6 \pm 7.4$} & \revised{$83.5 \pm 6.5$} & \revised{$86.6 \pm 6.1$} & \revised{$70.9 \pm 8.2$} \\
{\CheckmarkBold} & {\CheckmarkBold} & {\CheckmarkBold} & \revised{${88.6\pm6.7}$} &  \revised{$86.1\pm6.7$} & \revised{$88.0\pm10.1$} & \revised{$73.7\pm10.2$} \\
\hline
\end{tabular}}
\end{table}

\begin{table}[t]
\centering
\caption{Quantitative results of different uncertainty thresholds.}
\label{tab:thres}
\resizebox{1.0\columnwidth}{!}{
\begin{tabular}{l | c c c c | c c}
\hline
  & \multirow{2}{*}{AC (\%)} & \multirow{2}{*}{PR (\%)} & \multirow{2}{*}{RE (\%)} & \multirow{2}{*}{JA (\%)} & Labelling & Labelling \\
  &  &  &  &  &  Rate (\%) &  Accuracy (\%)\\
\hline
$\tau=\infty$ & $84.65$ & $86.34$ & $84.65$ & $70.43$ & $\bm{93.49}$ & $92.04$ \\
$\tau=0.2$ & $85.32$ & $\bm{86.71}$ & $85.06$ & $71.14$ & $92.72$ & $94.13$ \\
$\tau=0.1$  & $\bm{85.95}$ & $84.96$ & $\bm{87.05}$ & $\bm{71.43}$ & $87.51$ & $96.99$ \\
$\tau=0.05$ & $85.49$ & $86.19$ & $86.42$ & $71.20$ & $60.76$ & $\bm{99.04}$\\
\hline
\end{tabular}}
\end{table}

\begin{table}[t]
\centering
\caption{\revised{Comparison of labelling rate and labelling accuracy of pseudo labels generated by UATD in different iterations. ``TS" indicates the initial timestamp annotations.}}
\label{tab:iter}
{
\begin{tabular}{l | c c c c }
\hline
Iteration & TS & 1-st & 2-nd & 3-rd  \\
\hline
Labelling Rate (\%) & 0.33 & 67.70 & 76.82 & 84.45  \\
Labelling Accuracy (\%) & 100.00 & 98.69 & 97.95 & 97.42  \\
\hline
\end{tabular}}
\end{table}

\begin{figure}[t]
\centering
\includegraphics[width=0.48\textwidth,height=0.18\textheight]{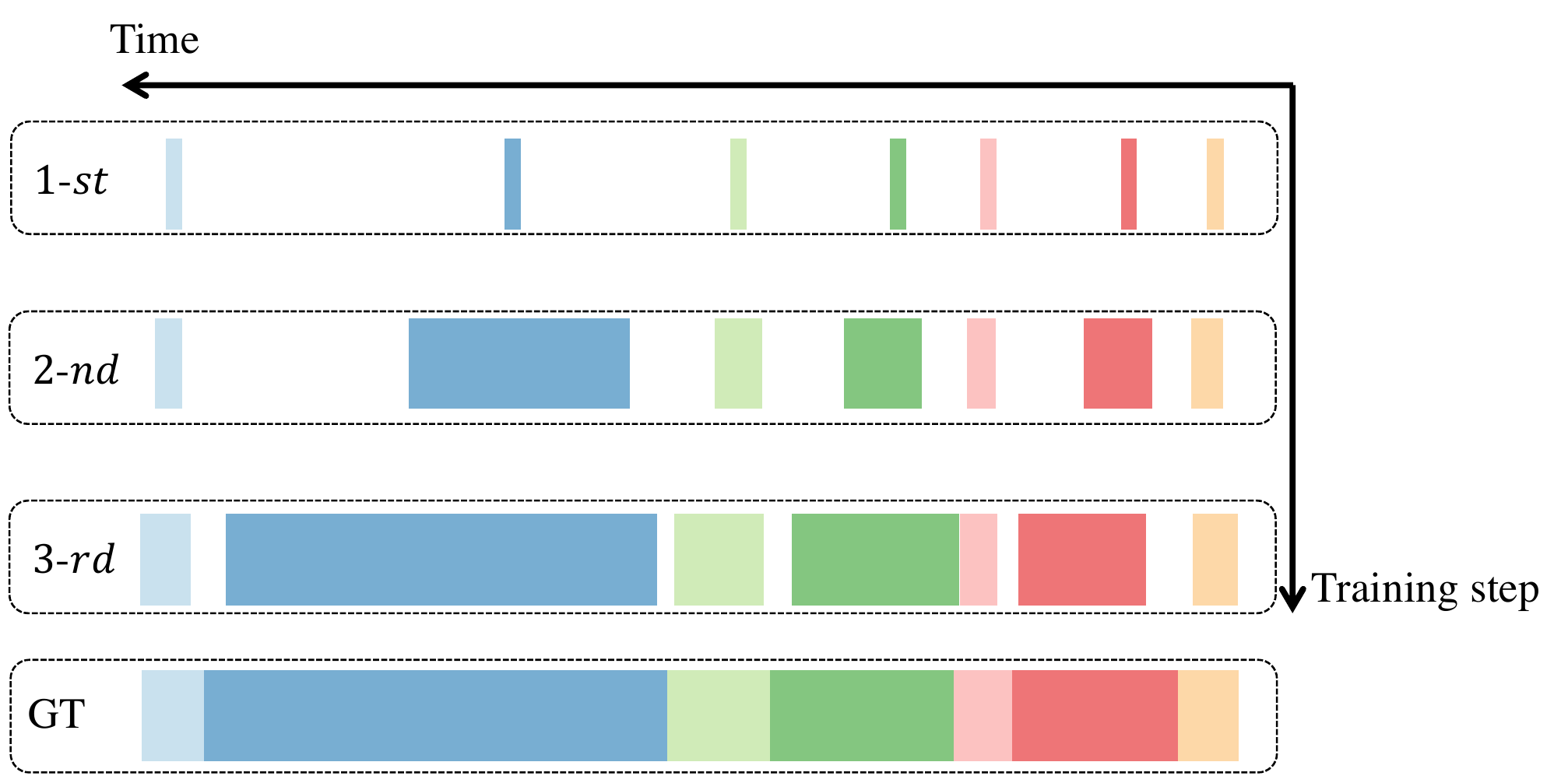}
\caption{
\revised{Visualization of the different iterations of the pseudo labels generated by our method. ``GT" indicates the ground truth. ``$1$-st", ``2-nd" and ``$3$-rd" indicate generated pseudo labels in the first, second and third iterations respectively.
}
}
\label{fig:temporaldiffusion}
\end{figure}

\begin{figure}[t]
\centering
\includegraphics[width=0.45\textwidth,height=0.1\textheight]{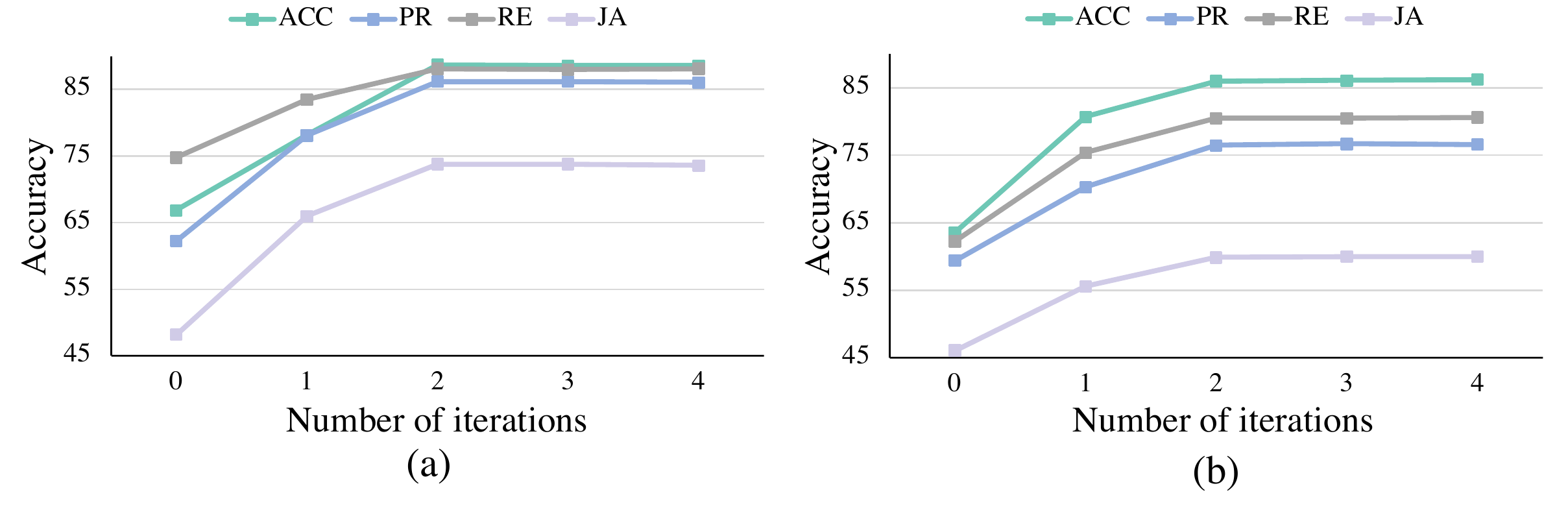}
\caption{
\revised{Analysis of the number of iterations for loop training on (a) Cholec80 and (b) M2CAI16.
We show the results of ACC, PR, RE and JA of models with different numbers of iterations.
}
}
\label{fig:numberloop}
\end{figure}

\begin{figure}[t]
\centering
\subfigure[Cholec80]{
\includegraphics[width=.22\textwidth]{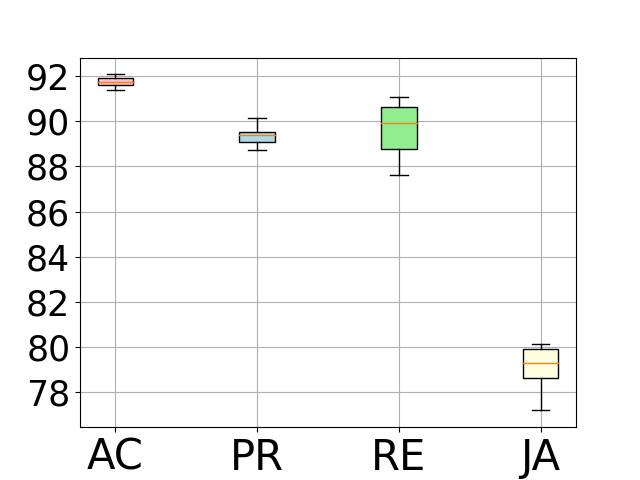}
}
\subfigure[M2CAI16]{
\includegraphics[width=.22\textwidth]{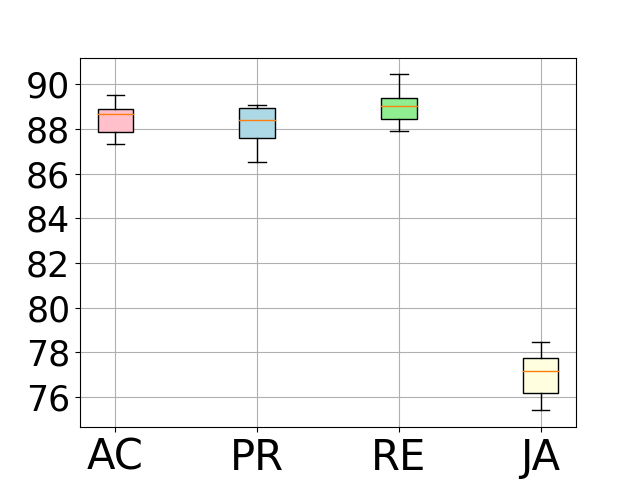}
}
\caption{Box plots of performance of random timestamp annotations on Cholec80 and M2CAI16 datasets.}
\label{fig:box}
\end{figure}

\begin{table}[t]
\centering
\caption{Quantitative results of Start, End, Middle and Random timestamp positions on Cholec80 dataset.}
\label{tab:extreme}
{
\begin{tabular}{l | c c c c }
\hline
Timestamp & \multirow{2}{*}{AC (\%)} & \multirow{2}{*}{PR (\%)} & \multirow{2}{*}{RE (\%)} & \multirow{2}{*}{JA (\%)} \\
Position &  &  &  &  \\
\hline
Start & $90.64$ & $87.92$ & $88.37$ & $76.75$ \\
End & $90.17$ & $88.35$ & $82.24$ & $70.75$ \\
Middle  & $\bm{92.59}$ & $\bm{90.13}$ & $\bm{89.60}$ & $\bm{80.04}$ \\
Random & $91.86$ & $89.51$ & $90.52$ & $79.90$ \\
\hline
\end{tabular}}
\end{table}

\begin{table}[t]
\centering
\caption{Comparison of annotation time between a single timestamp and two timestamps.}
\label{tab:singleandtwo}
{
\begin{tabular}{l | c c }
\hline
Video Index & 01 & 05 \\
\hline
Single Timestamp & 222s & 155s \\
Two Timestamps & 331s & 279s\\
\hline

\hline
\end{tabular}}
\end{table}
\begin{table}[t]
\centering
\caption{
\revised{Comparison of performance of models training with a single timestamp and two timestamps.}
}
\label{tab:onetworesults}
{
\resizebox{0.9\columnwidth}{!}{
\begin{tabular}{l | l  | c c c c}
\hline
Method & Annotation & AC (\%) & PR (\%) & RE (\%) & JA (\%) \\
\hline
\multicolumn{5}{c}{\revised{Cholec80}} \\
\hline
\multirow{3}{*}{\revised{Casual TCN}} 
 & \revised{Single}  & \revised{$88.56$} & \revised{$86.05$} & \revised{$88.00$} & \revised{$73.72$} \\
 & \revised{Two}  & $ \revised{88.79}$ & \revised{$89.61$}& $\revised{88.12}$ & $\revised{73.80}$ \\
\hdashline
\multirow{3}{*}{\revised{TCN}} 
 & \revised{Single} & \revised{$91.86$} & \revised{$89.51$} & \revised{$90.52$} & \revised{$79.90$} \\
 & \revised{Two} & $\revised{91.91}$ & $\revised{89.66}$ & $\revised{90.81}$ & $\revised{79.93}$ \\
\hline
\multicolumn{5}{c}{\revised{M2CAI16}} \\
\hline
\multirow{3}{*}{\revised{Casual TCN}}  & \revised{Single}  &  \revised{$86.03$} &  \revised{$85.02$} &  \revised{$87.08$} &  \revised{$71.43$} \\
 & \revised{Two}  & \revised{${86.07}$} & \revised{${85.11}$} & \revised{${87.14}$} & \revised{${71.50}$} \\
\hdashline
\multirow{3}{*}{\revised{TCN}} 
 & \revised{Single}  &  \revised{$87.62$} &  \revised{$88.25$} &  \revised{$87.91$} &  \revised{$75.72$} \\
 & \revised{Two}  & \revised{${87.70}$} & \revised{${88.30}$} & \revised{${87.98}$} & \revised{$\bm{75.81}$} \\
\hline
\end{tabular}}
}
\end{table}
\begin{figure}[t]
\centering
\includegraphics[width=0.45\textwidth,height=0.18\textheight]{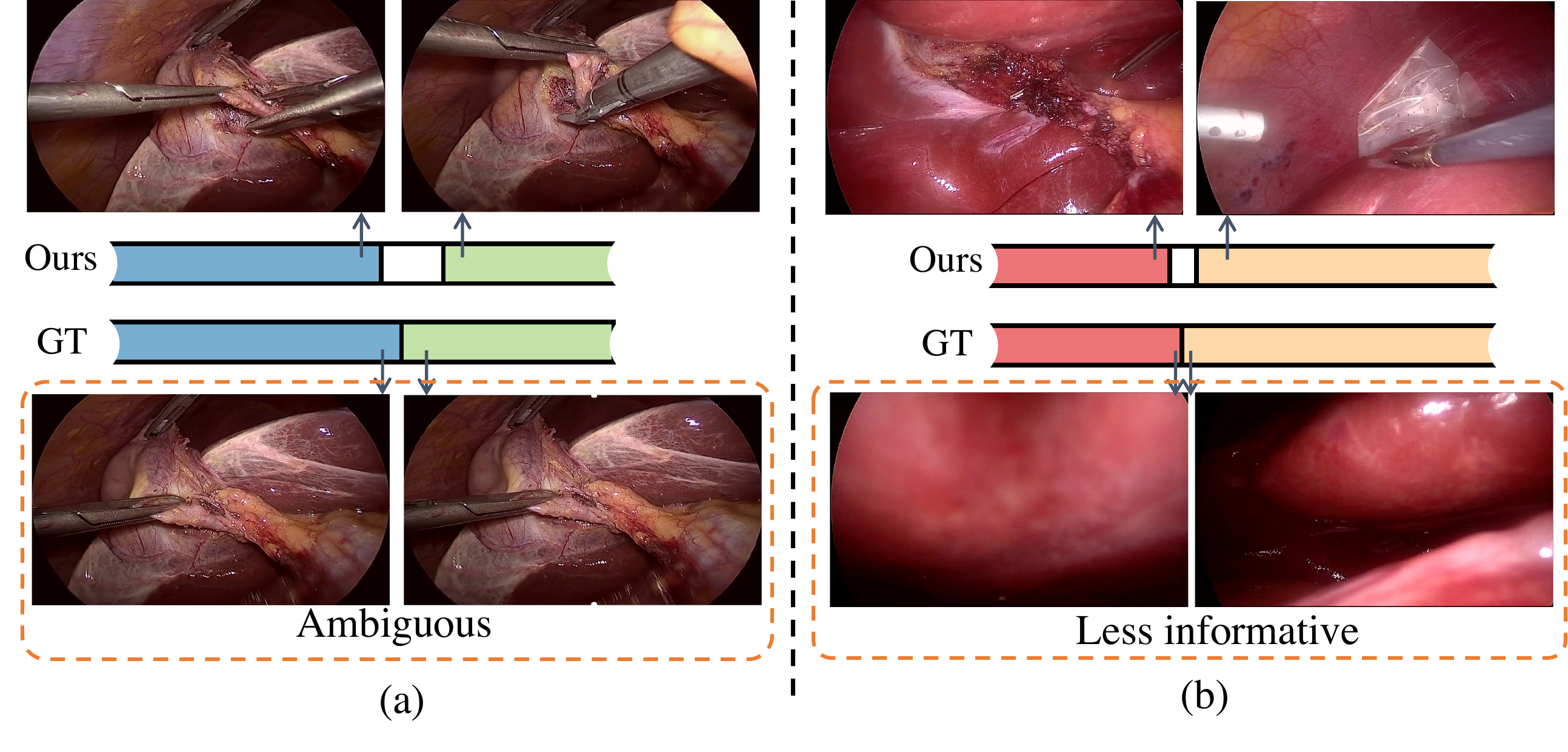}
\caption{
\revised{Comparison of pseudo labels generated by ours and ground-truth. It is clear that our method avoids annotating the frames near boundaries, where frames are generally (a) ambiguous or (b) less informative.
In our paper, we regard ambiguous frames as the frames that shows similar appearance in different phases following~\cite{ding2022exploring}.
Less informative frames indicate the frames that provide little information to identify different phases, such as phases containing no actions or instruments.
}
}
\label{fig:boundary}
\end{figure}

\subsection{Ablation Study} \label{sec:abla}

\noindent\textbf{Effect of UATD and LP.} There are two key components, \emph{i.e.}, uncertainty-aware temporal diffusion (UATD) and loop training (LP), in our method.
We ablate the effect of them in Table. \ref{tab:ablative}.
%
It is clear that the proposed UATD can improve the timestamp supervision with a clear margin, \emph{e.g.}, combined with UATD, the model achieves $85.59\%$ accuracy, outperforming $18.67\%$ over the baseline model.
Also, we could find that loop training contributes to around $3\%$ improvements.

\vspace{1.5mm}
\noindent \textbf{Impact of the uncertainty threshold~$\tau$.}
%
The quality of pseudo labels is depended on pseudo labeling rate and pseudo labels accuracy, which is controlled by the uncertainty threshold~$\tau$ in Algorithm~\ref{alg:diffsuion}.
%
In order to evaluate the effect of $\tau$, we compare the performance of the models with different $\tau$ and report the results in Table~\ref{tab:thres}.
\revised{
``Labelling Rate" indicates the ratio of the frames annotated by our method to all frames.
To evaluate the accuracy of our generated annotations,~\emph{i.e.}, pseudo labels, we compare the generated pseudo labels with the ground-truth.
Specifically, for a frame, if the annotated label generated by our method is equal to the ground-truth, the frame is regarded as the correct annotated frame, and vice versa. 
%
}
%
We can find that the higher uncertainty threshold would lead to the higher pseudo labeling rate and the lower accuracy of pseudo labels, and vice versa.
For example, with infinity threshold,~\emph{i.e.}, first row in Table~\ref{tab:thres}, pseudo labeling rate can reach $93.49\%$ while accuracy of pseudo labels is only $92.04\%$.
Such higher labeling rate would introduce more noisy labels, which degrades the labeling accuracy.
\revised{Furthermore, with different $\tau$,~\emph{i.e.}, $0.2$, $0.1$ and $0.05$, the performance of our method is very stable. For example, the variance for accuracy values with different thresholds is only $0.11\%$}
In our paper, we set $\tau$ to $0.1$ for the best trade-off.

\vspace{1.5mm}
\noindent \revised{\textbf{Analysis of pseudo labels in different iterations.}}
{
Given only a single manual labeled annotations, we show that our model can generate more and more reliable pseudo-labels step by step in Table~\ref{tab:iter}.
%
``Labelling Rate" and ``Labelling Accuracy" are the same meaning as Table~\ref{tab:thres}.
It shows that our method can generate more and more pseudo labels the number of iterations increases.
This is because each iteration of temporal diffusion gives temporal model extra information, the model can generate more pseudo labels next time.
Also, the accuracy of generated pseudo labels is very trustworthy.
Since the frames show very similar appearances to their adjacent frames, the network can easily generate correct predictions for the neighbor frames of the annotated frame.
We also show the visualization of the \revised{different iterations of the pseudo labels generated by our method in Fig.~\ref{fig:temporaldiffusion}.}
}

\vspace{1.5mm}
\noindent \revised{\textbf{Effect the number of iterations for loop training.}}
\revised{
We conduct the analysis of the number of iterations for loop training in Fig.~\ref{fig:numberloop}.
The results show that more iterations for loop training can improve the performance, since more trustworthy labels are introduced to training.
We also find that there is no significant performance improvement after more than two iterations.
Hence, in this paper, we set the number of iterations for loop training as two.
}

\vspace{1.5mm}
\revised{\noindent \textbf{Robust to different timestamp annotations.}}
~\label{sec:robust}
In our experiments, the timestamp annotations are generated by randomly selecting one frames to be annotated of each phase.
\revised{In order to evaluate our method is robust to the different timestamps, we random $10$ different timestamps by different random seeds and analyse their impacts to the performance, which is shown in Fig.~\ref{fig:box}.
Specifically, we report the box plots of $10$ random timestamp annotations on Cholec80 and M2CAI16 datasets.
}
%
%
%
The short and flat boxes indicate that our proposed method is robust to different timestamp annotations,\revised{~\emph{e.g.}, the difference between the maximum and minimum is $2.3\%$.}
What's more, our method can outperform most of methods in Table.~\ref{tab:sota} with even the worst timestamp annotations.

\vspace{1.5mm}
\revised{\noindent \textbf{Effect of timestamps in different phase positions.}}
~\label{sec:different_positions}
\revised{
In order to explore the effect of timestamps in different phase positions, we enforce the random timestamp annotations inside start, end or middle region of each phase.
More specifically, we regard the first 10\% frames, the middle 10\% frames and the last 10\% frames of each phase as the start, middle and end regions.
}
As shown in Table.~\ref{tab:extreme}, annotating at the start and end frames of each phase would degrade the performance.
This is because that frames near boundaries are generally ambiguous, which can be hard to act as an anchor of temporal diffusion.
In the contrast, the middle frames are more discriminative to represent current phases and thus can generate more correct pseudo labels.
Actually, the surgeons~,\emph{i.e.}, the annotators, tend to label the discriminative frames because they can easily recognize them when seeing through the whole video~\cite{ma2020sf}, which ensure timestamp annotations efficient and effective.
%

\vspace{1.5mm}
\revised{\noindent \textbf{Comparison between a single and two timestamps.}}
~\label{sec:comparison_single_two}
During the timestamp annotation, once a phase is identified and current timestamp is recorded, the surgeon could choose to record another timestamp for the phase.
Here, we compare the annotation cost between a single and two timestamps in Table~\ref{tab:singleandtwo}.
Two videos,~\emph{i.e.},``01" and ``05", are sampled from Cholec80 and M2CAI16 respectively.
The result shows that two timestamp annotation would cost more time than a single timestamp annotation,~\emph{e.g.}, the surgeon would spend $331$s for ``01" while annotating a single timestamp only requires $222$s. 
We also conduct experiments to compare the performance of the models training with a single timestamp and two timestamps, as shown in Table~\ref{tab:onetworesults}. The results show that two timestamp annotation cannot achieve clear improvement but bring additional annotation cost. Hence, annotating a single timestamp is much efficient than two timestamps, and we use the best efficient way to solve surgical phase recognition in this paper. 

\vspace{1.5mm}
\noindent \textbf{Comparison of generated pseudo label and ground-truth.}\label{sec:TvsF}
In our experiments, we find that our method only generates pseudo labels for discriminative frames while ignores the ambiguous ones near boundaries.
%
As shown in Fig.~\ref{fig:boundary}, our generated pseudo labels discard ambiguous or \revised{less informative} frames compared to the ground-truth.
More importantly, the model trained with our generated pseudo labels outperforms the model trained with the ground-truth; see details in Table. \ref{tab:sota}.
%
%
\revised{This indicates that ambiguous boundary of two adjacent actions would degree the performance.}
\begin{figure}[t]
\centering
\subfigure[Ours]{
\includegraphics[width=.22\textwidth]{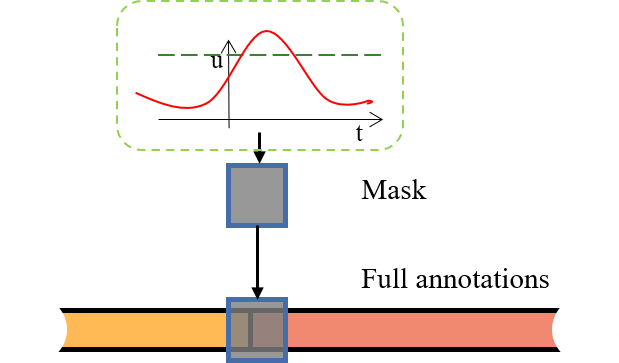}
}
\subfigure[Fixed width]{
\includegraphics[width=.22\textwidth]{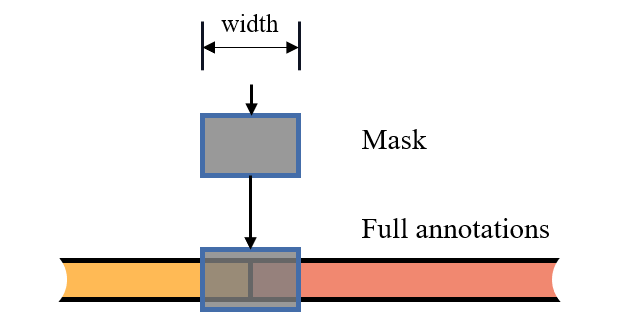}
}
\caption{(a) Masking boundaries by using UATD to detect ambiguous frames. (b) Masking boundaries by fixed width.}
\label{fig:mask}
\end{figure}


\begin{table}[t]
\centering
\caption{
\revised{Effectiveness of incorporating UATD into current methods.} `Timestamp' is using our generated pseudo labels by UATD from timestamp annotations and `GT w/ UATD' indicates the ground-truth labels masked by UATD; see Sec.~\ref{sec:maskinggt} for details.
}
\label{tab:ours_full}
{
\resizebox{0.9\columnwidth}{!}{
\begin{tabular}{l | l  | c c c c}
\hline
Method & Annotation & AC (\%) & PR (\%) & RE (\%) & JA (\%) \\
\hline
\multicolumn{5}{c}{\revised{Cholec80}} \\
\hline
\multirow{3}{*}{Casual TCN} & GT   & $87.94$ & $86.40$ & $84.81$ & $72.40$ \\
 & Timestamp  & $88.56$ & $86.05$ & $88.00$ & $73.72$ \\
 & GT w/ UATD  & $\bm{91.18}$ & $\bm{89.88}$ & $\bm{90.93}$ & $\bm{79.76}$ \\
\hdashline
\multirow{3}{*}{TCN} & GT  & $91.14$ & $90.84$ & $87.64$ & $79.14$ \\
 & Timestamp & $91.86$ & $89.51$ & $90.52$ & $79.90$ \\
 & GT w/ UATD  & $\bm{92.75}$ & $\bm{91.23}$ & $\bm{93.10}$ & $\bm{83.89}$ \\
\hline
\multicolumn{5}{c}{\revised{M2CAI16}} \\
\hline
\multirow{3}{*}{\revised{Casual TCN}} & \revised{GT}   & \revised{$81.91$} & \revised{$84.82$} & \revised{$82.24$} & \revised{$68.06$} \\
 & \revised{Timestamp}  &  \revised{$86.03$} &  \revised{$85.02$} &  \revised{$87.08$} &  \revised{$71.43$} \\
 & \revised{GT w/ UATD}  & \revised{$\bm{87.01}$} & \revised{$\bm{88.23}$} & \revised{$\bm{88.81}$} & \revised{$\bm{76.26}$} \\
\hdashline
\multirow{3}{*}{\revised{TCN}} &  \revised{GT}   & \revised{$82.94$} & \revised{$85.82$} & \revised{$82.69$} & \revised{$69.71$} \\
 & \revised{Timestamp}  &  \revised{$87.62$} &  \revised{$88.25$} &  \revised{$87.91$} &  \revised{$75.72$} \\
 & \revised{GT w/ UATD}  & \revised{$\bm{88.32}$} & \revised{$\bm{89.03}$} & \revised{$\bm{89.23}$} & \revised{$\bm{78.81}$} \\
\hline
\multicolumn{5}{c}{\revised{JIGSAW}} \\
\hline
\multirow{3}{*}{\revised{Casual TCN}} & \revised{GT}   & \revised{$80.12$} & \revised{$82.02$} & \revised{$81.16$} & \revised{$69.11$} \\
 & \revised{Timestamp}  &  \revised{$81.73$} &  \revised{$84.91$} &  \revised{$84.82$} &  \revised{$71.55$} \\
 & \revised{GT w/ UATD}  & \revised{$\bm{83.27}$} & \revised{$\bm{85.51}$} & \revised{$\bm{85.27}$} & \revised{$\bm{72.81}$} \\
\hdashline
\multirow{3}{*}{\revised{TCN}} &  \revised{GT}   & \revised{$81.43$} & \revised{$84.29$} & \revised{$83.71$} & \revised{$70.18$} \\
 & \revised{Timestamp}  &  \revised{$83.18$} &  \revised{$85.19$} &  \revised{$85.72$} &  \revised{$72.12$} \\
 & \revised{GT w/ UATD}  & \revised{$\bm{84.28}$} & \revised{$\bm{86.13}$} & \revised{$\bm{86.16}$} & \revised{$\bm{73.15}$} \\
\hline
\end{tabular}}
}
\end{table}

\begin{table}[t]
\centering
\caption{Effectiveness of boundary mask on Cholec80 dataset.}
\label{tab:boundary}
{
\begin{tabular}{l c c c c}
\hline
Mask width & AC (\%) & PR (\%) & RE (\%) & JA (\%) \\
\hline
0 & $91.14$ & $90.84$ & $87.64$ & $79.14$ \\
3 & $92.04$ & $91.87$ & $89.07$ & $81.44$ \\
5 & $92.31$ & $92.26$ & $89.52$ & $82.12$ \\
10 & $\bm{92.75}$ & $92.86$ & $\bm{90.57}$ & $\bm{83.32}$ \\
20 & $92.68$ & $\bm{93.20}$ & $90.40$ & $82.66$ \\
\hline
\end{tabular}}
\end{table}
\begin{figure}[t]
\centering
\includegraphics[width=0.5\textwidth,height=0.3\textheight]{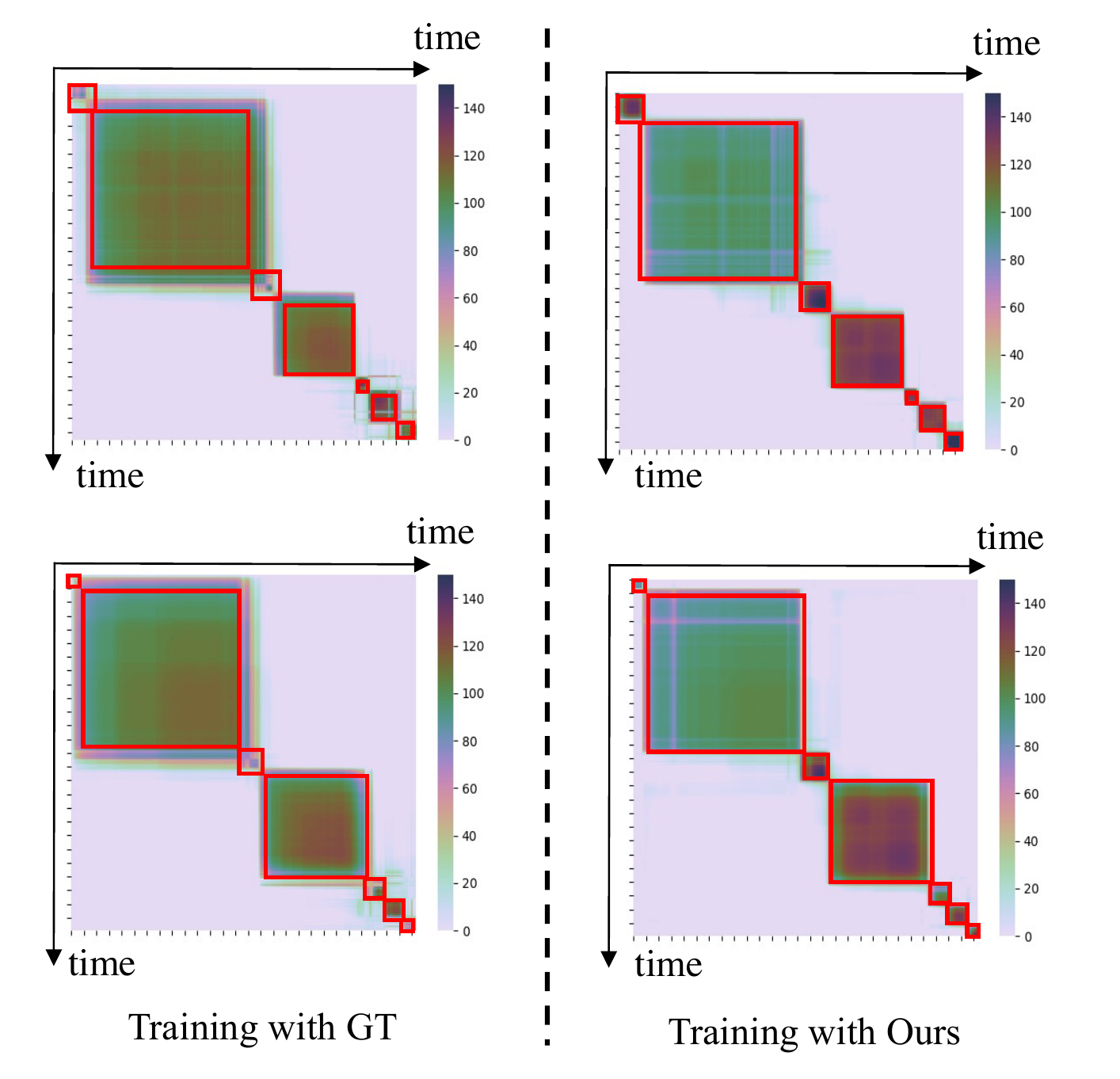}
\caption{
\revised{Feature similarity matrix visualization. The horizontal and vertical axes represent the time indexes. We use cosine similarity to measure the degree of similarity between two arbitrary frame-level feature vectors within the same video. Each red box indicates each phase in a video. Note that, frame-level features of the same phase should be as similar as possible while separating one from others. Compared to the model trained with the ground-truth (GT), better representations of the features can be learned by our generated pseudo labels (right).}
}
\label{fig:sim}
\end{figure}
\subsection{Incorporate UATD into Current Methods}~\label{sec:maskinggt}
As analyzed in Fig.~\ref{fig:boundary}, we find that our method can only generate labels for discriminative frames, instead of ambiguous frames.
\revised{It comes up that if masking ambiguous frames from the ground-truth by our UATD can improve the performance.}
%
To this end, as shown in Fig.~\ref{fig:mask}~(a), we mask some ground-truth labels near boundaries, based on the pseudo labels generated by our methods.
%
%
To be specific, we use UATD to generate pseudo labels, and record the indexes of unlabelled frames that are with high uncertainty.
Then, we remove those frames with the recorded indexes from the ground-truth, and obtained a clean ground-truth to supervise the model.
%
%
%
%
We compare the performance of the models training with \textbf{(a)} \revised{ground-truth (GT)}, \textbf{(b)} pseudo labels generated by UATD and \revised{\textbf{(c)}} GT masked by UATD, and report the results in Table~\ref{tab:ours_full}.
\revised{To show the generalization of our proposed method, we also conduct experiments on JIGSAW~\cite{ahmidi2017dataset}, which a simulated dataset with a clear domain gap.}
\revised{From Table~\ref{tab:ours_full}}, it is clear that the model training with GT masked by UATD can achieve the best results, even outperforms the current SOTA; see Table~\ref{tab:sota} for comparison.
%
%
%
We further conduct experiments on the models training with GT masked by the fixed width, as shown in Fig.~\ref{fig:mask}~(b).
As illustrated in Table.~\ref{tab:boundary}, masking some frames near boundary during training outperform the model without masking over around $1\%-3\%$ in all metrics.
However, it will introduce a new hyper-parameter,~\emph{i.e.}, the width of mask, which is critical to the performance.
Hence, in order to achieve the good performance, we need to conduct many experiments to find the best choice, which is very time-consuming.
%
%
\emph{On the contrary, our method can be used as an approach to clean the noisy labels in the ground-truth automatically, without the need of hand-designed width.}
{To further explain this phenomenon, we visualize the feature similarity matrix in Fig.~\ref{fig:sim}.}
{Each red box in each similarity matrix indicates each phase in a video.}
It is clear that training with our generate pseudo labels~\emph{i.e.}, removing ambiguous labels near boundaries between two phases, would help to decrease intra-class distance and increase inter-class distance simultaneously.

%
%


\section{Discussion}
Surgical phase recognition is one key component of computer-assisted surgery systems, which advances context-awareness in modern operating rooms.
However, most existing works require full annotations which are expensive, expertise-required and error-prone~\cite{dipietro2019automated}.
In contrast, we introduce timestamp supervision which only requires one timestamp annotated by human for each phase in a video.
{We invite two surgeons to conduct both full and timestamp annotations and record the time cost for these two annotations.}
{To leverage this supervision, we propose \textbf{U}ncertainty-\textbf{A}ware \textbf{T}emporal \textbf{D}iffusion (\textbf{UATD}) to generate trustworthy pseudo labels for those unlabeled frames, which is based on the property of surgical phases.}
Furthermore, loop training is also introduce to address the imbalance training and memory cost in timestamp surgical phase recognition.
%
The in-depth empirical studies of the proposed UATD and LP based on timestamp supervision discovers four deep insights: \textbf{1)} Timestamp annotation can reduce $74\%$ annotation time compared with the full annotation, and surgeons tend to annotate those timestamps that are near the middle of phases;
\textbf{2)} Extensive experiments demonstrate that our method can achieve competitive results compared with full supervision methods, while reducing manual annotation cost;
\textbf{3)} Less is more in surgical phase recognition,~\emph{i.e.}, less but discriminative pseudo labels outperform full but containing ambiguous frames;
\textbf{4)} The proposed UATD can be used as a plug and play method to clean ambiguous labels near boundaries between phases, and improve the performance of the current surgical phase recognition methods; see details in Table~\ref{tab:ours_full}.

Although our method achieves promising results, there are some limitations.
First, the temporal property we consider is not overall yet.
The diffusion in our method assumes that the workflow is smooth without dramatic change and hardly any ambiguous frame occurs in the internal of phase~\revised{; see Fig.~\ref{fig:vis} (e)-(f)}.
But such assumption may be false for other datasets and in the future we will study more comprehensive temporal relationship to handle the intra-phase discontinuity.
Moreover, the training process we propose is time-consuming containing several iterations of training model from scratch.
And we will design more elegant training process to link up the optimal learning from different annotations,~\emph{i.e.}, different rounds of temporal diffusion in our methods.

Finally, we expect \revised{the} community to focus more on label-efficient surgical video analysis. The weakly setting of videos, such as transcripts~\cite{huang2016connectionist} and timestamp supervision, deserve further attention and exploiting. And the related ideas can be further investigated in other medical image analysis problems in CT~\cite{li2018h,gibson2018automatic,heimann2009comparison, li2020transformation}, MRI~\cite{wang2020ica,li20183d,yu2020use}.

\section{Conclusion}
In this paper, we introduce the most annotation-saving setting, namely timestamp supervision, for surgical phase recognition. 
With timestamp supervision, we propose a novel uncertainty-aware temporal diffusion (UATD) method to generate trustworthy pseudo labels according to the labeled frames. 
Our main idea is to generate pseudo labels by considering the relationship among video frames.
Results on two datasets show that our method can achieve the competitive performance compared with the fully supervised setup.
Moreover, we also find that our method can be used as a labeling clean approach to remove the noisy labels near boundaries to improve the generalization of the current surgical phase recognition, which reveals an interesting phenomenon less is more in this task.
This paper provides some insights for label-efficient surgical phase recognition and hopefully inspire researchers to design label-efficient surgical video analysis algorithms.

\bibliographystyle{ieeetr}
\small\bibliography{tmi}

\end{document}